\documentclass[runningheads]{llncs}
\usepackage{graphicx}  
\usepackage{makecell}  
\usepackage{booktabs}  

\usepackage{eccv}

\usepackage{subcaption}
\usepackage{eccvabbrv}

\usepackage{graphicx}
\usepackage{booktabs}

\usepackage[accsupp]{axessibility}  

\usepackage{orcidlink}
\usepackage{float}

\begin{document}

\title{\textbf{MV-GEL}: Language-Driven Multi-View Geometric Entity Localization on Meshes} 

\titlerunning{\textbf{MV-GEL}: Multi-View Geometric Entity Localization on Meshes}

\author{Kartik Bali\inst{1,2}\orcidlink{0009-0005-3474-5830} \and
Roland Aydin\inst{3}\orcidlink{0000-0002-9542-9146}}

\authorrunning{K.~Bali et al.}

\institute{Helmholtz Zentrum Hereon, Max-Planck-Strasse 1, 21502 Geesthacht, Germany \and
Institute for Continuum and Material Mechanics, Hamburg
University of Technology, Eissendorfer Strasse 42, 21073
Hamburg, Germany \and German Research Center for Artificial Intelligence,
Campus D3 2, 66123 Saarbrücken, Germany\\
\email{kartik.bali@hereon.de}, \email{roland.aydin@dfki.de}\\}

\maketitle

\begin{abstract}

 Identifying and grounding precise geometric entities, such as edges, planar regions, and curved surfaces within 3D objects, is foundational to computer-aided design (CAD), robotic manipulation, and scientific simulation. Although modern Vision Language Models (VLMs) have advanced referring segmentation (RIS) in the image domain, extending such language-driven localization to structured 3D geometry is substantially harder. The 3D object appearance is highly sensitive to viewpoints; a single perspective may render a target entity clearly observable, while another may suffer from severe occlusion or foreshortening. In this work, we attempt to solve these challenges with \textbf{MV-GEL} (Multi-View Geometric Entity Localization), a framework for localizing fine-grained geometric entities on polygon meshes from natural language queries. Our key insight is that reliable CAD entity (i.e., faces, edges or solids) localization depends on selecting views that make the queried entity maximally interpretable. We introduce \textit{GELviews}, a prompt-conditioned ranking module that prioritizes viewpoints based on language prompted observability of geometric CAD entities. Selected views are processed by a VLM-based reasoning segmentation backbone, and predicted masks are lifted to the corresponding meshes via geometry-aware ray casting. Our framework is completely CAD agnostic and relies only on 3D meshes. Experiments show up to a 1.7× improvement in face-level IoU and over 4.5× gains in edge-level F1 compared to vanilla baselines, substantially outperforming CLIP-based and random view sampling, particularly for thin and view-sensitive structures. The dataset, code and trained checkpoints are available at \url{https://github.com/kbali1297/MV-GEL}. 

\end{abstract}

\section{Introduction}
\label{sec:intro}

Three-dimensional semantic localization is a fundamental prerequisite for intelligent systems to perceive, reason, and act in real-world environments. While classical approaches emphasize global pose estimation or reconstruction, modern applications demand precise identification of semantically meaningful object parts. In robotics, grasp planning depends on functional regions such as handles; in CAD and manufacturing, simulation and editing require selecting exact geometric entities including fillets, chamfers, and load-bearing surfaces. These targets are defined as discrete topological elements present in a 3D structure and their spatial identification is crucial for the relevant task.

\begin{figure}[t]
\centering
\includegraphics[width=1.03\columnwidth]{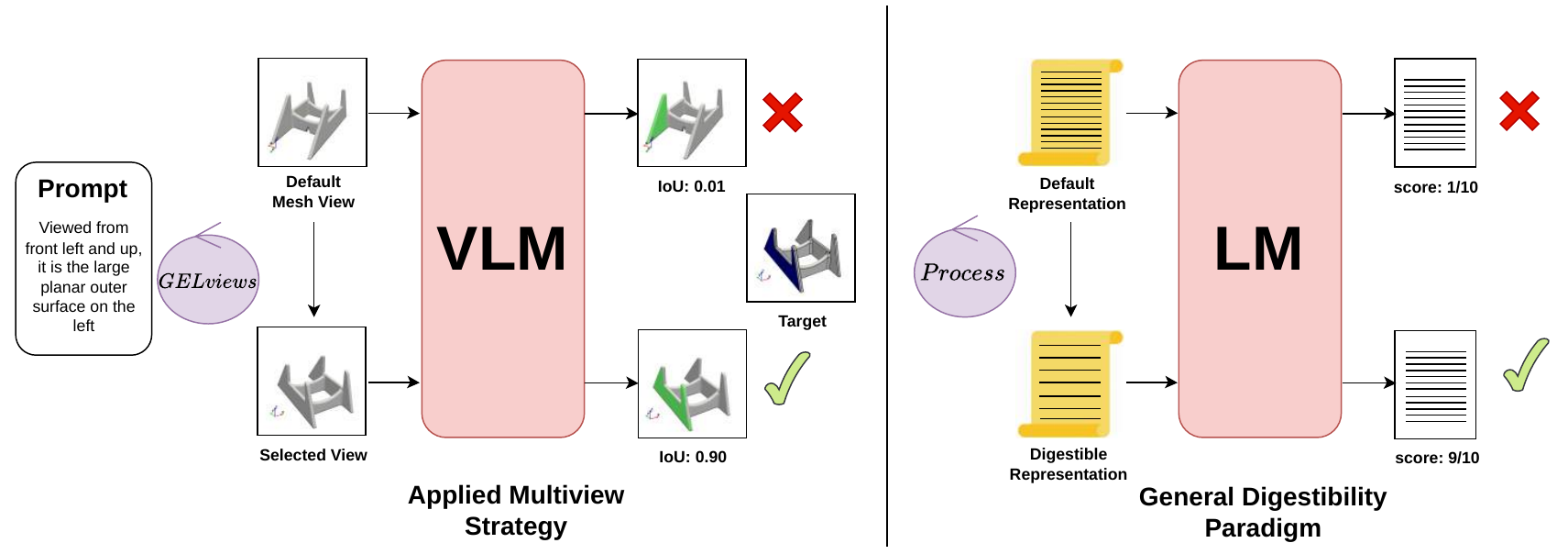}
\caption{We can tie our multi-view strategy to a more general digestibility paradigm, where certain representations of data are more interpretable to a language model than others}
\label{fig:Digestibility_paradigm}
\end{figure}

Recent vision-language models have demonstrated strong semantic reasoning in images \cite{liu2023visual, bai2023qwen, chen2024internvl, team2024gemini}, and a few have explored reasoning-based segmentation \cite{lai2024lisa, ren2024pixellm, yuan2024osprey}. However, using these frameworks for 3D segmentation and entity localization depends critically on how visual information is presented to the VLM. In 3D settings, the same underlying geometry can appear either highly informative or severely ambiguous depending on the viewpoint. Self-occlusion, depth conflation, and thin structural elements often render relevant entities visually indigestible from arbitrary projections. Thus, language-driven geometric localization is not merely a segmentation problem, but a question of making structured 3D information interpretable to current VLMs that process visual information in 2D.

In this work, we address this visual interpretability problem for VLMs in the context of entity localization for mesh objects, derived from CAD. We introduce a prompt-conditioned ranking framework that estimates entity-specific geometric visibility across candidate views, presenting the reasoning segmentation model with geometrically digestible evidence that helps it achieve accurate entity localization, as illustrated in Fig.  \ref{fig:Digestibility_paradigm}. 
In summary, our contributions are threefold:
\begin{itemize}
    \item We create a dataset and formalize the task of locating geometric entities in CAD meshes guided via natural language. 
    \item We propose \textit{GELviews}, a prompt-conditioned view ranking framework that prioritizes viewpoints given to reasoning segmentation 2D VLMs, based on entity-specific geometric visibility.
    \item We introduce a CAD-agnostic framework \textbf{MV-GEL} for localizing these geometric entities using text prompts directly on the mesh.
\end{itemize}

We evaluate the results of our framework by introducing a mesh-aware lifting strategy that maps 2D segmentation masks to discrete topological mesh entities (mesh faces and edges).

\section{Related Work}

\subsection{Vision-Language Models and Referring Segmentation}
The integration of visual and textual modalities has driven remarkable progress in open-vocabulary image understanding. Foundational vision-language models (VLMs) like CLIP \cite{radford2021learning}, ALIGN \cite{jia2021scaling}, and BLIP-2 \cite{li2023blip} align image and text embeddings, enabling zero-shot recognition. Building on this, Large Multimodal Models (LMMs) such as LLaVA \cite{liu2023visual}, InstructBLIP \cite{dai2023instructblip}, and Qwen-VL \cite{bai2023qwen} have demonstrated sophisticated visual reasoning capabilities.

In parallel, referring expression segmentation (RES) has evolved from relying on complex multi-modal fusion networks \cite{yu2018mattnet, ding2021vision, wang2022cris, liu2023gres} to leveraging foundation models like the Segment Anything Model (SAM) \cite{kirillov2023segment, zhao2023fast, zou2023segment}. Successors like LISA \cite{lai2024lisa}, PixelLM \cite{ren2024pixellm}, Osprey \cite{yuan2024osprey}, and GSVA \cite{xia2024gsva} further improve semantic segmentation via pixel-level dense reasoning using Vision Language Models. However, these models are trained predominantly on natural images (e.g., COCO, LVIS) and struggle to comprehend the rigid geometry, thin structures, and geometric description cues inherent to Computer-Aided Design (CAD). Our work bridges this domain gap by explicitly fine-tuning a reasoning VLM on CAD mesh renderings and geometry in natural language used to spatially describe these features.

\subsection{Language Guided Segmentation in 3D}

Analyzing 3D geometry via 2D projections is a highly effective paradigm, popularized by MVCNN \cite{su2015multi} and expanded by recent view-graph and transformer architectures \cite{wang2019dominant, goyal2021revisiting}. Recently, the field has shifted toward lifting 2D VLM features into 3D space to achieve open-vocabulary 3D understanding. Techniques such as ConceptFusion \cite{jatavallabhula2023conceptfusion}, OpenScene \cite{peng2023openscene}, and LERF \cite{kerr2023lerf} project CLIP features onto point clouds or Neural Radiance Fields (NeRFs). For part-level segmentation, PartSLIP \cite{liu2023partslip}, PartSLIP++ \cite{zhou2023partslip++}, and SATR \cite{abdelreheem2023satr} leverage 2D grounding cues and GLIP \cite{li2022grounded} to localize semantic parts in 3D point clouds. PointCLIP \cite{zhang2022pointclip} and PointCLIPV2 \cite{zhu2023pointclip} adapt vision-language models to point clouds through multi-view rendering and feature alignment, while COPS \cite{garosi20253d} and GeoZE \cite{mei2024geometrically} further improve open-vocabulary understanding via geometry-aware aggregation of 2D visual features and geometric priors. More recent approaches, such as PARIS3D \cite{kareem2024paris3d}, PointBind \cite{guo2023point}, ReferSplat \cite{he2025refersplat}, and IPDN \cite{chen2025ipdn}, incorporate multimodal reasoning, structural prompts, and multi-view constraints for language-guided 3D localization. Recent foundation-model-based methods such as Find3D \cite{ma2025find} and PatchAlign3D \cite{hadgi2026patchalign3d} learn language-aligned point- and patch-level representations for open-vocabulary part retrieval and segmentation. While these methods achieve impressive results on unstructured point clouds, Gaussian splats, and scene-level 3D representations, adapting them to fine-grained mesh entities with explicit topological connectivity remains challenging.

\subsection{Deep Learning on CAD and Boundary Representations}
Unlike standard meshes or point clouds, CAD models are defined by Boundary Representations (B-Reps), i.e graphs of parametric curves and surfaces. Early geometric deep learning adapted to B-Reps through UV-Net \cite{jayaraman2021uv}, BRepNet \cite{lambourne2021brepnet}, and SB-GCN \cite{jones2021automate} for tasks like solid classification and machining feature recognition, heavily utilizing the ABC Dataset \cite{koch2019abc} and Fusion 360 Gallery \cite{willis2021fusion}.

Recently, text-driven 3D generation has extended into the CAD domain. Works like Text2CAD \cite{khan2024text2cad}, CAD-LLaMA \cite{li2025cad}, and SolidGen \cite{jayaraman2022solidgen} generate CAD construction sequences from text. Furthermore, B-repLer \cite{liu2025b} and CAD-MLLM \cite{xu2024cad} introduced semantic editing of B-Reps using language models. Despite these advances, the fine-grained localization of arbitrary, language-described CAD entities (e.g., ``the inner chamfered edge'') remains largely unexplored. We address this by providing a unified pipeline for language-driven geometric entity localization on meshes.

\subsection{Prompt-Aware View Selection}
In multi-view pipelines, selecting optimal viewpoints is critical for information gain. While traditional Next-Best-View (NBV) algorithms rely on geometric heuristics \cite{connolly1985determination} or deep reinforcement learning \cite{chen2024gennbv}, the advent of VLMs has shifted the focus toward semantic view selection. Methods like Cap3D \cite{luo2023scalable} and ViewRefer \cite{guo2023viewrefer} dynamically rank camera poses by maximizing global CLIP similarity between 2D renderings and a language prompt. However, global semantic matching fails on homogeneous, textureless industrial CAD parts, where pinpointing fine-grained features (e.g., ``the inner chamfered edge'') requires explicit geometric awareness. Furthermore, while recent 3D referring segmentation methods tackle localized spatial understanding, they predominantly operate on unstructured point clouds or voxels, and are not designed to output exact topological entities defined by a mesh graph.  

Our work bridges these disconnected domains. 
To the best of our knowledge, MV-GEL is the first framework to address prompt-conditioned, multi-view localization of exact topological entities on 3D meshes.

\section{Problem Formulation}

Let $\mathcal{M} = (V, \mathcal{E}, \mathcal{F})$ denote a discrete 3D triangle mesh, where $V = \{\textbf{v}_i \in \mathbb{R}^3\}_{i=1}^{N}$ is the set of vertices, $\mathcal{E} = \{e_k\}_{k=1}^{K}$ is the set of edges, and $\mathcal{F} = \{f_j\}_{j=1}^{M}$ is the set of triangular faces. In the context of computer-aided design, a \textit{geometric entity} refers to a distinct topological element, for instance, a continuous surface patch or a boundary curve.

Given a natural language query $q$ describing a specific geometric entity (e.g., ``the inner cylindrical surface'', ``the outer chamfered edge''), our goal is to localize the exact subset of mesh elements corresponding to this queried feature. Crucially, while the query conceptually describes an analytical CAD entity (a native B-Rep face or edge), the system is provided only the unstructured polygon mesh $\mathcal{M}$ as its geometric input at inference.

Formally, let $\mathcal{X} \in \{\mathcal{E}, \mathcal{F}\}$ represent the target domain of the queried entity type (edges or faces). We formulate geometric entity localization as predicting a binary labeling function $\hat{y} : \mathcal{X} \rightarrow \{0,1\}$, 
where $\hat{y}(x_i) = 1$ indicates that the discrete mesh element $x_i \in \mathcal{X}$ belongs to the target structure described by $q$. The objective is to produce a topologically consistent prediction $\hat{y}$ over $\mathcal{M}$ such that the selected discrete mesh elements accurately reconstruct the continuous geometric entity intended by the user.

\section{Dataset Construction}


We construct a mesh–language localization dataset derived from the ABC CAD dataset. 
We sample entities and generate over 54k natural query-geometric entity pairs from over 5,000 CAD geometries spanning diverse mechanical parts 
with varying topology and surface complexity. For each CAD model, we randomly sample a single boundary representation (B-rep) primitive, 
restricted to either one surface (face) or one edge. As highlighted in the previous section, the sampled CAD entities provide the target views and segmentation maps during training, given the mesh $\mathcal{M}$ and the input query $q$.

\subsubsection{Multi-View Rendering and Ranking}

For each CAD mesh $\mathcal{M}$, we generate a discrete set of candidate camera poses $\mathcal{V} = \{v_i\}$ distributed across uniform azimuth and elevation intervals. To evaluate a pose $v_i$ with respect to a target geometric entity $\mathcal{E}$ (an edge or a face), we first select a representative subset of points $\mathcal{P}_\mathcal{E} \subset \mathcal{E}$ using Farthest Point Sampling (FPS). We estimate the geometric observability of $\mathcal{E}$ from pose $v_i$ by casting rays from the camera center through the virtual image plane toward each point $p_j \in \mathcal{P}_\mathcal{E}$. A point is visible if the ray intersection is unobstructed by the mesh surface. We define the visibility score as $S(v_i, \mathcal{E}) = \sum_{p_j \in \mathcal{P}_\mathcal{E}} \mathbf{1}\{p_j \text{ is visible from } v_i\}$.

Because our visibility testing is computed from the camera's perspective, views where the entity is viewed at a slanted or grazing angle are naturally penalized. In such configurations, rays are more likely to encounter self-occlusion or grazing intersection errors, leading to a reduced count of visible points in $S(v_i, \mathcal{E})$. Consequently, our ranking mechanism inherently prioritizes "head-on" viewpoints that provide clear, unobstructed geometric exposure of the entity. Finally, for the top-ranked poses, we generate two synchronized renderings: an unannotated RGB image for model inference and a secondary mask-highlighted rendering for supervised training.

\subsubsection{Language Query Construction}

For each sampled entity's maximal view, we construct a natural language query that uniquely identifies its geometric entity using only intrinsic geometric and topological cues, similar to how humans describe geometric entities. The descriptions rely primarily on properties such as curvature, orientation, adjacency, symmetry, continuity, and relative position within the part. This formulation enforces geometry-aware and viewpoint-specific grounding. Detailed data generation prompts can be found in the supplementary.

\begin{figure}[t]
\centering
\includegraphics[width=1\columnwidth]{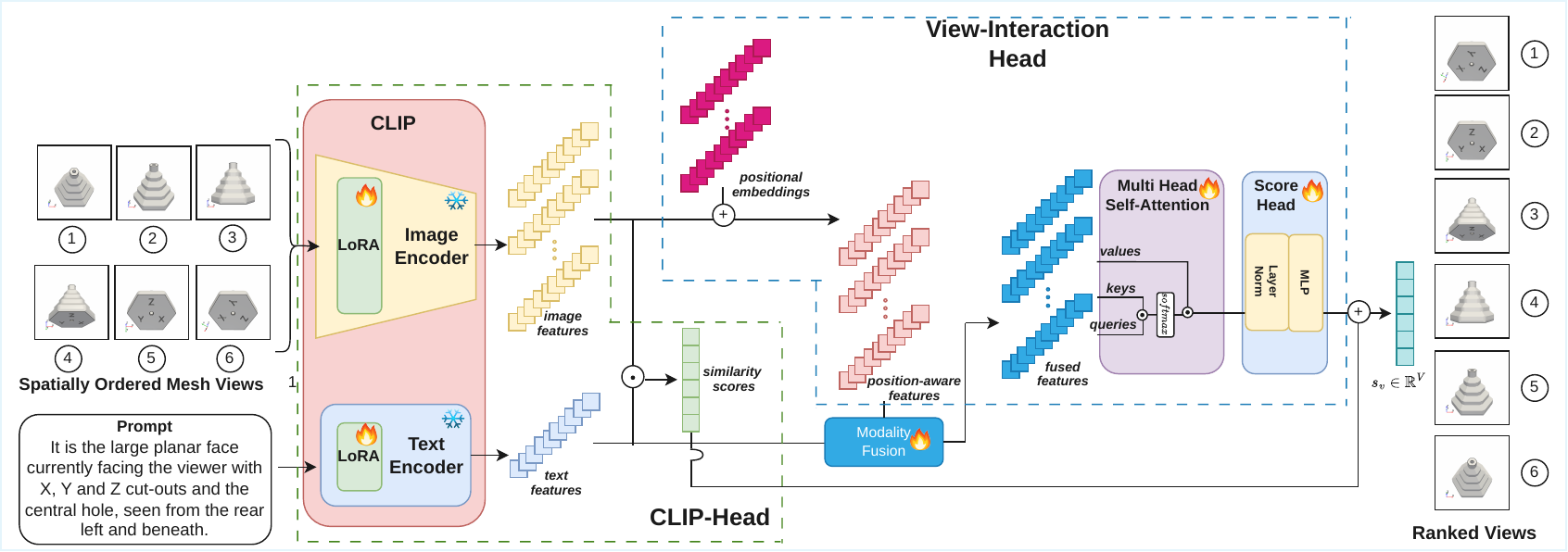}
\caption{\textit{GELviews}: Our View selection framework takes the natural language prompt containing the geometric and view-point descriptions and returns semantically optimal views. In this example, the views with planar face at the bottom with cut-outs get higher ranks than the top views.}
\label{fig:GELviews}
\end{figure}

\section{Method}

Given a triangle mesh $\mathcal{M}$ and a natural language query $q$ describing a geometric structure, our objective is to localize the subset of mesh topological entities corresponding to the queried entity. 



\subsection{Prompt-Conditioned Geometric View Ranking (\textit{GELviews})}

We formulate view selection as a prompt-conditioned geometric ranking problem. Given a set of candidate viewpoints, the objective is to prioritize views that maximize geometric observability of the queried entity. Unlike global semantic alignment approaches, this formulation requires reasoning about structural visibility, occlusion, and entity-specific spatial exposure across views.

\subsubsection{CLIP Head}

Each rendered image $I_v$ is encoded using a CLIP vision encoder 
$\mathcal{E}_{img}$, and the query $q$ is encoded using the CLIP 
text encoder $\mathcal{E}_{txt}$: $f_v =\mathcal{E}_{img}(I_v),t = \mathcal{E}_{txt}(q)$,
where $f_v, t \in \mathbb{R}^{D}$ are the global aligned feature embeddings (CLS token embeddings). 
These embeddings represent high-level semantic summaries of the view 
and the query, respectively, and are aligned due to the CLIP training objective.
We noticed that base CLIP image and text embeddings are not optimized to contain geometric semantic information and yield poor view selection when obtained from the base CLIP encoders.
To mitigate this, both encoders are adapted using low-rank parameter updates, allowing 
task-specific specialization while preserving the pretrained 
vision–language alignment prior. This ensures stable optimization 
and prevents catastrophic forgetting of CLIP’s semantic structure. As shown in the Fig. \ref{fig:GELviews}, for an initial estimate of view relevance, we compute normalized cosine similarity 
between each view embedding and the query embedding: $s_v^{\text{clip}} = 
    \left\langle 
    \frac{f_v}{\|f_v\|}, 
    \frac{t}{\|t\|}
    \right\rangle$.



\subsubsection{View Positional Encoding}

To incorporate geometric context, specifically to make the network aware of view-points and their geometric order, we inject learnable positional embeddings 
$p_v \in \mathbb{R}^{D}$ encoding the identity and relative ordering 
of each camera pose: $\tilde{f}_v = f_v + p_v$.



\subsubsection{Modality Fusion}

While $s_v^{\text{clip}}$ measures independent alignment, 
view ranking requires conditioning image features explicitly 
on the query semantics. We therefore apply a fusion operator 
$\mathcal{F}$ between the view features and the global text embedding:$\hat{f}_v = \mathcal{F}(\tilde{f}_v, t)$.
We investigate two complementary conditioning mechanisms: 
(i) FiLM \cite{perez2018film}, which performs feature-wise affine modulation $\text{FiLM}(\tilde{f}_v, t) = \gamma(t) \odot \tilde{f}_v + \beta(t)$, 
and (ii) Cross-Attention \cite{vaswani2017attention, lu2019vilbert}, 
which enables token-level interaction between view features and the text embedding. FiLM provides lightweight global semantic conditioning and 
strictly generalizes additive and weighted-add fusion schemes 
as special cases (via learned scaling $\gamma(t)$ and bias $\beta(t)$), 
while preserving the pretrained CLIP representation at initialization. 
In contrast, Cross-Attention enables more expressive multimodal interaction 
by allowing query-dependent reweighting across feature dimensions. 


\subsubsection{View Interaction Head}

Individual view scoring ignores the fact that geometric visibility 
is inherently relative across viewpoints. For example, a surface 
may only be fully visible in one view while partially occluded in others. 
To capture such dependencies, we process the fused features using 
a multi-layer Transformer encoder $\mathcal{T}$ operating along 
the view dimension: $h_v = \mathcal{T}(\{\hat{f}_v\}_{v=1}^{V})_v$. 
This self-attention mechanism allows each view to attend to all 
other views, enabling context-aware comparison and geometric reasoning. 
As a result, the representation of each view becomes informed by 
the global distribution of viewpoints. A lightweight scoring head maps the context-aware features 
to scalar geometric adjustment scores: $s_v^{\text{geo}} = 
    \mathbf{w}^\top \mathrm{LN}(h_v)$,
where $\mathrm{LN}$ denotes LayerNorm. This component learns 
to correct or refine the initial semantic similarity based on 
geometric exposure and cross-view reasoning.
The final ranking score combines semantic alignment and 
geometry-aware adjustment $s_v = \alpha \left(s_v^{\text{clip}} + s_v^{\text{geo}}\right)$, 
where $\alpha$ is a learnable temperature parameter controlling 
score sharpness. Views are ranked according to $s_v$, and the 
top-$K$ views are selected greedily from the \textit{GELviews} scores.

\subsubsection{Pairwise Ranking Objective}

\textit{GELviews} is trained to prioritize views that best expose the queried geometric entity. 
For each sample, we compute a visibility score $r_v$ per view and define 
$\mathcal{V}^+$ as the top-$k$ views, where 
$k = \max(1, \lfloor V \cdot \rho \rfloor)$ with a small fraction $\rho$. 
All remaining views form $\mathcal{V}^-$. 
This reduces ranking to separating geometrically strong views from weaker ones. Given predicted scores $s_v$, we enforce a margin $m$ between positive and negative views. 
For all $(i,j)$ with $i \in \mathcal{V}^+$ and $j \in \mathcal{V}^-$, we minimize
\begin{equation}
    \mathcal{L}_{rank} =
    \frac{1}{|\mathcal{P}|}
    \sum_{(i,j) \in \mathcal{P}}
    \log \left(
    1 + \exp\left(
    - (s_i - s_j - m)
    \right)
    \right),
\end{equation}
where $\mathcal{P}$ is the set of positive–negative pairs. 
This soft margin ($m=0.1$) objective encourages separation of informative viewpoints 
while avoiding noisy supervision over fine-grained ordering among weak views.

\subsubsection{View Non-Maximum Suppression}

The ranked views produced by \textit{GELviews} may contain angularly redundant viewpoints that observe nearly identical surface regions. In our experiments we realised that applying view-NMS, contrary to popular practices, hurts our metrics. This is because views nearby act as fallback for entity segmentation (especially in cases where VLM might miss them on the main view) and thus aid in recalling the entities as views monotonically increase. As a result we have omitted it from our main analysis. Readers are advised to have a look at Table \ref{tab:baselines}, where the ViewNMS version for top-3 views (for angle threshold 4$5^\circ$) performs much worse than the regular greedy selection.  

\begin{figure}[t]
\centering
\includegraphics[width=1\columnwidth]{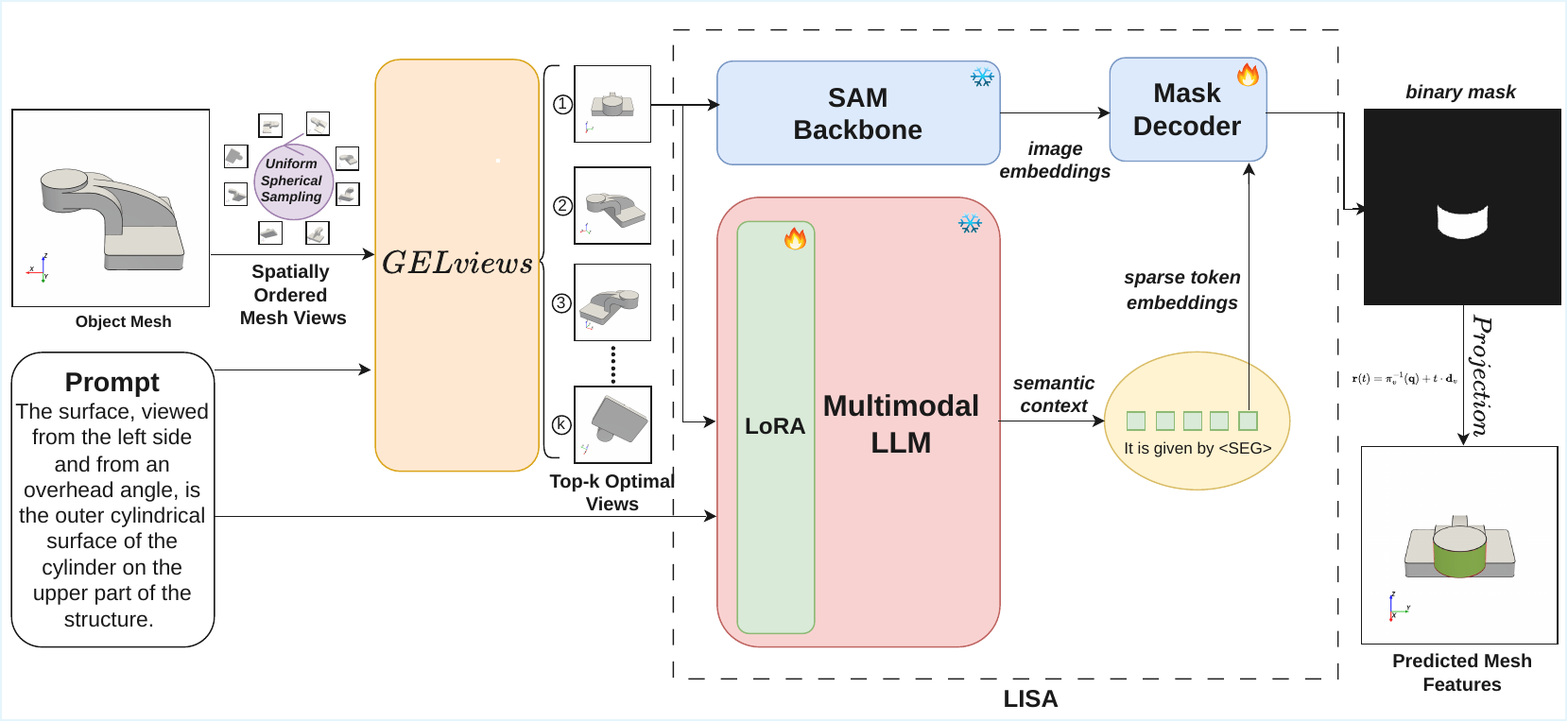}
\caption{\textit{MV-GEL Pipeline}: In our framework, we use \textit{GELviews} to compute the top-k views most relevant to localize a feature mentioned in the prompt. These views are then given sequentially to LISA-CAD for semantic segmentation to obtain view binary masks. Finally these binary masks are projected to the final mesh to obtain mesh features (edges/faces).}
\label{fig:MV-GEL}
\end{figure}

\subsection{Language-Driven Segmentation with LISA}

We adopt LISA \cite{lai2024lisa} as the underlying language-conditioned segmentation backbone. The model operates in image space and predicts a binary mask conditioned on a special \texttt{<SEG>} token, whose hidden representation is projected into the SAM mask decoder as a sparse prompt embedding. While segmentation and training supervision is performed on 2D ground truth masks, evaluation is defined on CAD mesh entities via a geometry-aware lifting mechanism.

We retain the original LISA training objectives, including the autoregressive language modeling loss and mask supervision (binary cross-entropy and DICE). 

\subsection{Modular Multi-View Localization Pipeline}

We now summarize the full MV-GEL pipeline, in Fig. \ref{fig:MV-GEL}, in functional form. 
Given a mesh $\mathcal{M}$ and query $q$, we first render a set of calibrated views $I_k = \mathcal{R}(\mathcal{M}, \pi_k), \quad k = 1,\dots,V$,
where $\pi_k$ denotes camera parameters and $\mathcal{R}$ is a rendering operator.

\paragraph{View Selection.}
The prompt-conditioned ranking module \textit{GELviews} selects a subset of informative views $\mathcal{S} = \Psi(\{I_k\}_{k=1}^{V}, q)$,
where $\mathcal{S} \subset \{1,\dots,V\}$ contains the top-$K$ geometrically relevant viewpoints.

\paragraph{Image-Space Segmentation.}
For each selected view $k \in \mathcal{S}$, the segmentation backbone predicts a binary mask $m_k = \Gamma(I_k, q)$, 
where $\Gamma$ denotes the LISA-based segmentation model operating in image space.

\paragraph{Geometric Lifting.}
The predicted masks are lifted to mesh topology via ray projection, producing a 3D labeling $\hat{y} = \mathcal{L}(\{m_k\}_{k \in \mathcal{S}}, \{\pi_k\}_{k \in \mathcal{S}}, \mathcal{M})$, 
where $\mathcal{L}$ denotes the geometry-aware lifting operator mapping foreground pixels to mesh faces or edges.
The overall localization is therefore obtained by composition $\hat{y} = \mathcal{L} \circ \Gamma \circ \Psi (\mathcal{M}, q)$,
with \textit{GELviews} ($\Psi$) and the segmentation backbone ($\Gamma$) trained independently. 
This modular design preserves pretrained segmentation priors and avoids joint optimization through the rendering process.




\begin{figure}[t]
    \centering
    \setlength{\abovecaptionskip}{2pt}
    \setlength{\belowcaptionskip}{0pt}
    \setlength{\textfloatsep}{6pt}
    \captionsetup[subfigure]{font=footnotesize, skip=2pt}

    \begin{subfigure}[t]{0.49\linewidth}
        \centering
        \includegraphics[height=2.9cm]{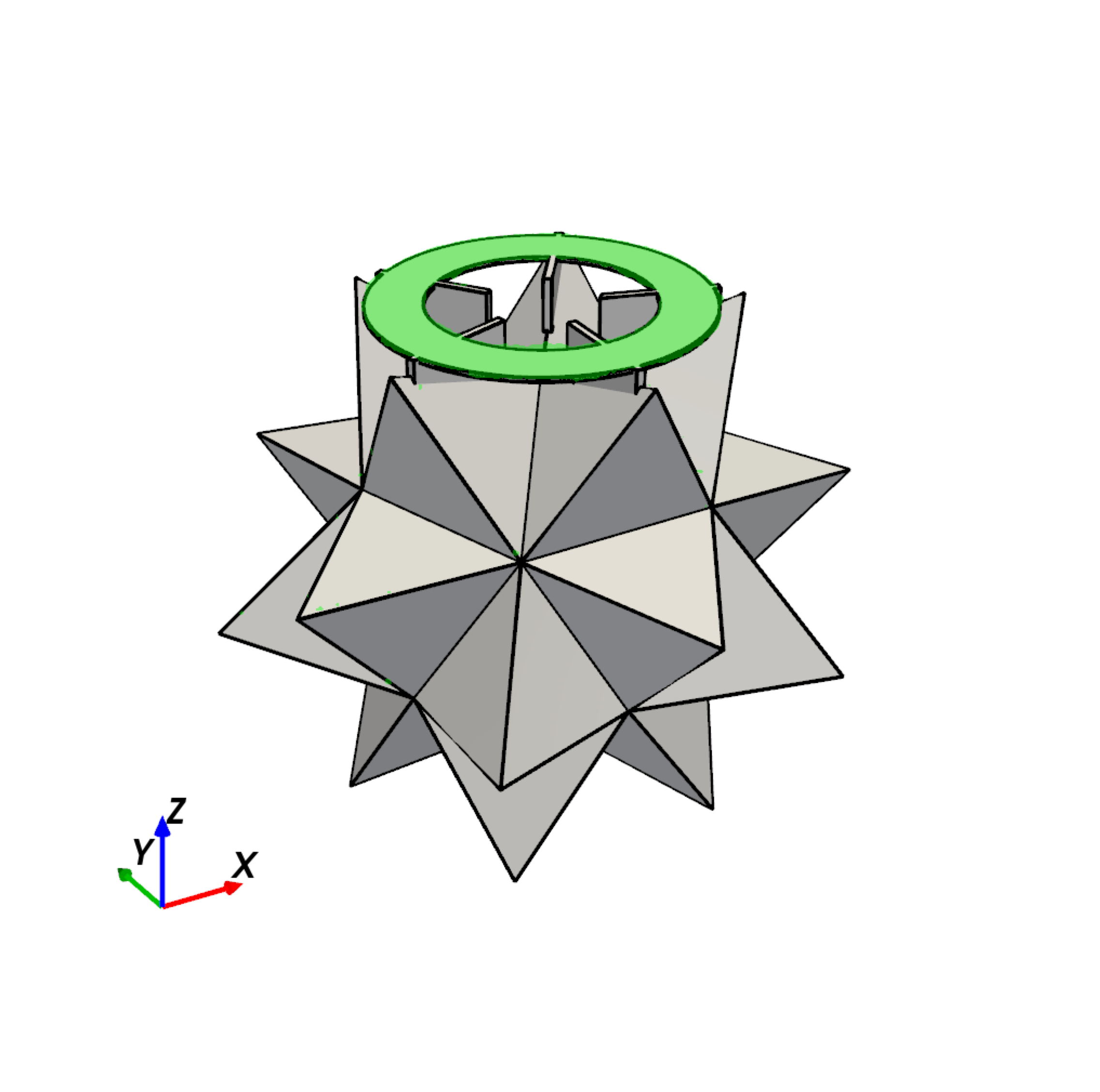}
        \caption{Looking at it from the front-left corner and a higher vantage point, the surface is the top annular ring face with top axis parallel to z-axis surrounding the circular hole.}
        \label{fig:top_left}
    \end{subfigure}
    \hfill
    \begin{subfigure}[t]{0.49\linewidth}
        \centering
        \includegraphics[height=2.9cm]{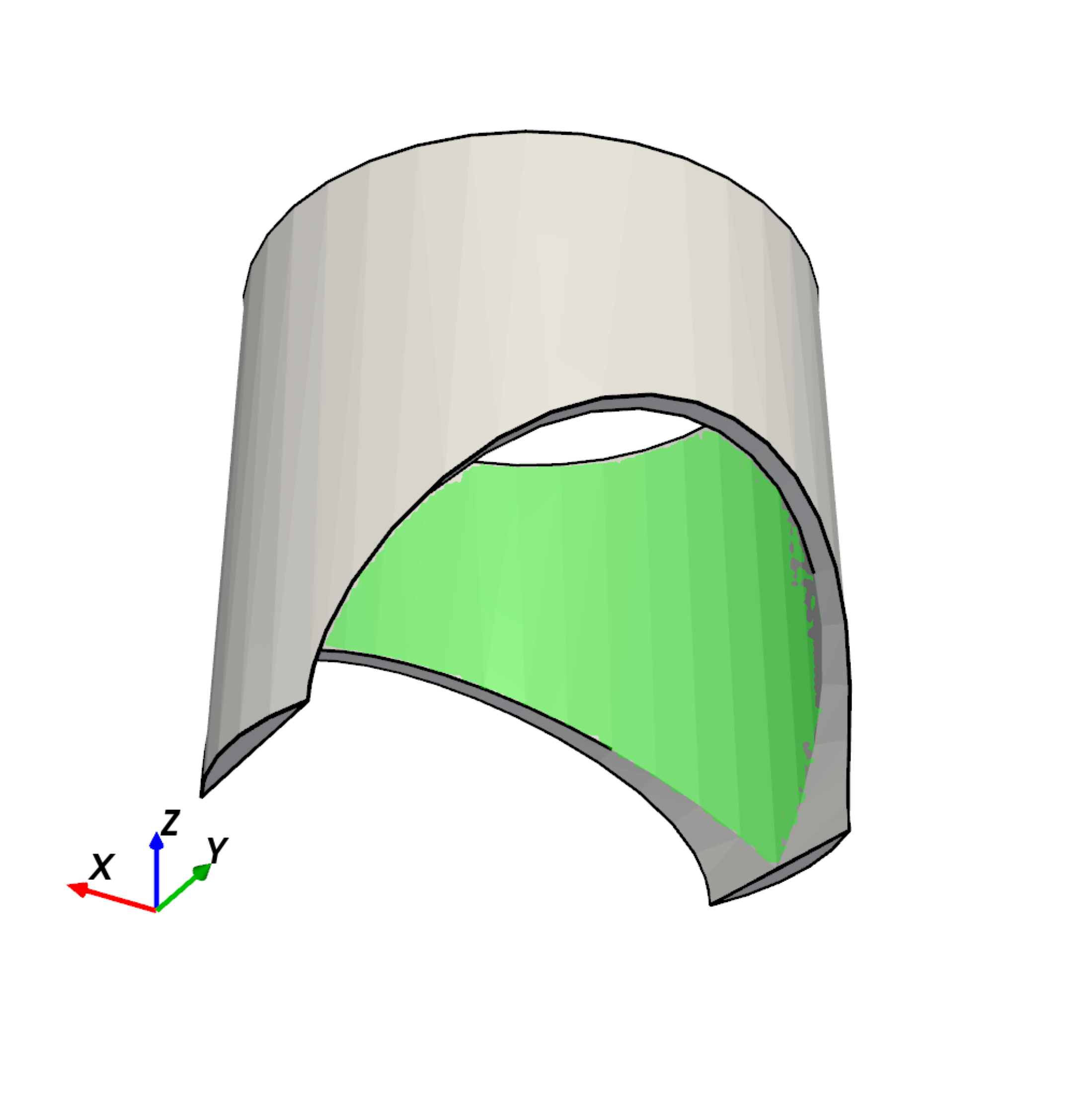}
        \caption{It is the inner surface of the twisted band-continuous concave interior surface extending from the upper inner rim down to the lower pointed inner opening, most clearly seen from front-left looking up at it.}
        \label{fig:top_right}
    \end{subfigure}

    \vspace{-2mm}

    \begin{subfigure}[t]{0.49\linewidth}
        \centering
        \includegraphics[height=2.9cm]{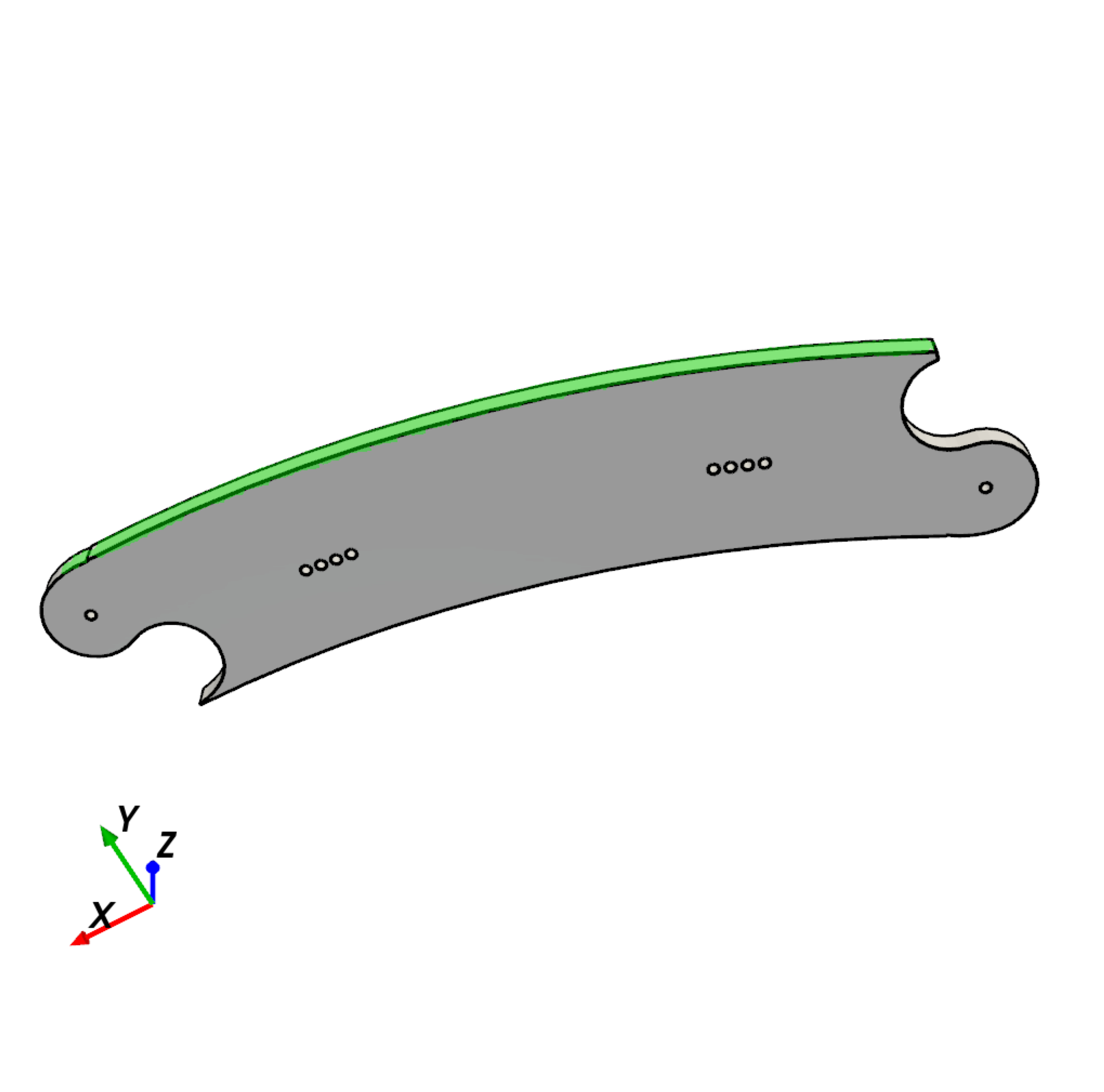}
        \caption{The edge, with viewpoint shifted towards rear-left and viewed from below, is the outer long convex longitudinal edge of the plate.}
        \label{fig:bot_left}
    \end{subfigure}
    \hfill
    \begin{subfigure}[t]{0.49\linewidth}
        \centering
        \includegraphics[height=2.9cm]{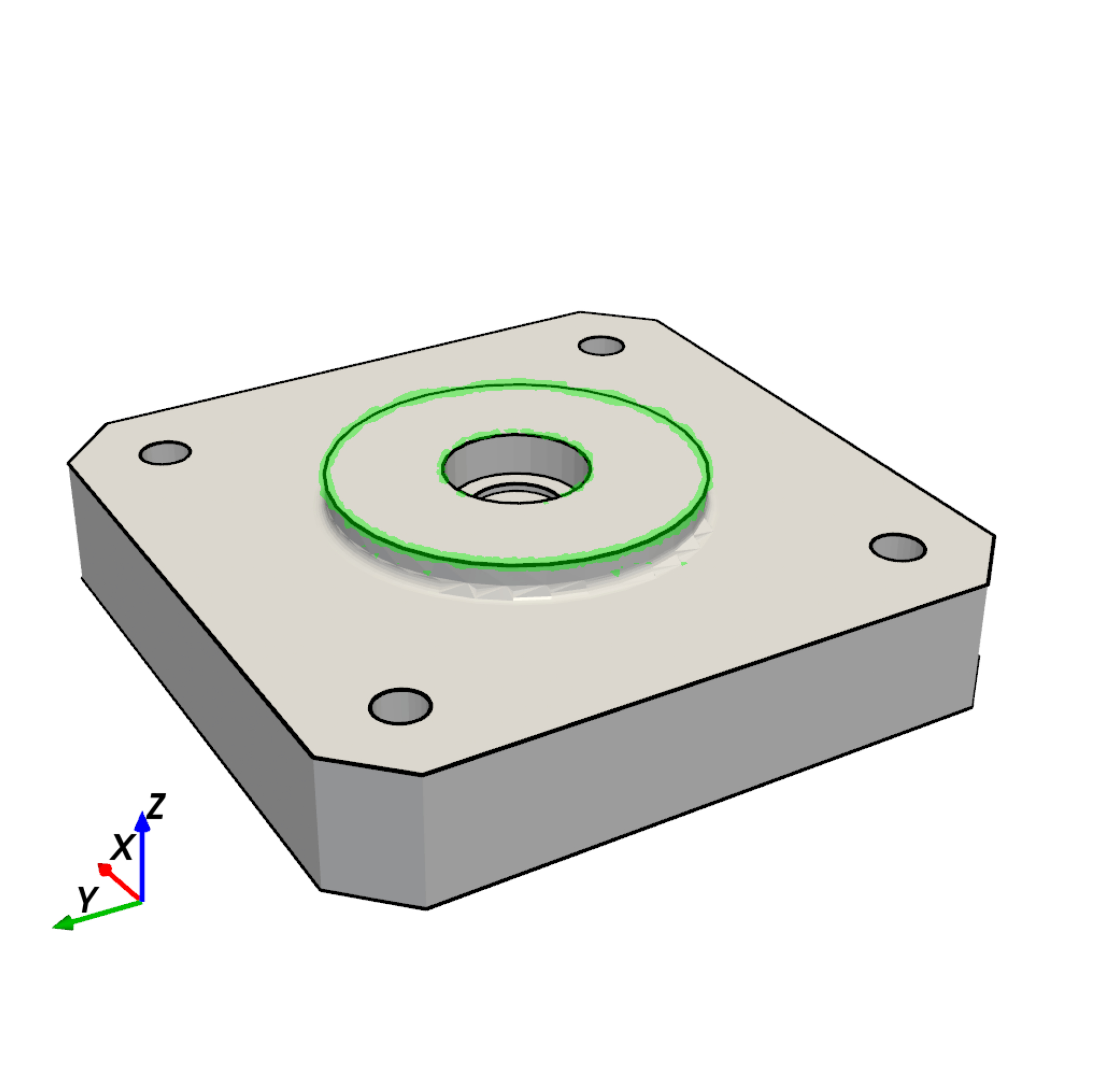}
        \caption{From the left side of the rear face and from a superior vertical position, its the circular edge at the base of the raised boss where it meets the top surface of the plate.}
        \label{fig:bot_right}
    \end{subfigure}

    \vspace{2mm}
    \caption{\footnotesize Examples of language-driven face and edge localization. The green mask denotes entity segmentation on the meshes via our \textbf{MV-GEL} framework.}
    \label{fig:qualitative_examples}
\end{figure}

\section{Experiments}


\subsubsection{Training}

We train \textit{GELviews} on 54k language queries paired with annotated top-performing views. LoRA adapters were used to fine tune clip image and text encoders with r=8 and alpha=16, with a layer dropout equal to 0.1. For each query, we uniformly sample candidate viewpoints over elevations in $[-60^\circ, 60^\circ]$ and azimuths in $[0^\circ, 360^\circ)$ at $30^\circ$ increments, yielding 60 candidate views per instance. The ranking module is optimized to prioritize views that maximize entity-specific geometric observability. We place the top 5 ($\rho = 0.09$) views in the $\mathcal{V}^+$ set and the rest in $\mathcal{V}^-$. 
For segmentation, we fine-tune \texttt{LLaVA-7B} in the LISA-CAD setting using the same 54k query–binary mask pairs. Training is conducted on 4 NVIDIA H100 GPUs, updating only the LoRA adapter layers and the mask decoder while keeping the remaining backbone parameters frozen. We use a learning rate of $1\times10^{-4}$ for all our trainable modules. 
Evaluation is performed on a held-out test set of 1535 query–entity pairs spanning 218 CAD meshes. After topology-aware lifting, performance is reported using face and edge-level metrics computed directly on the CAD mesh.

\subsubsection{Evaluation Strategy}

We evaluate localization directly on mesh entities rather than in image space. Predictions and ground truth are represented as sets of faces or edges, and overlap is computed using geometric weighting to reduce variations due to mesh discretization. Let $\mathcal{S}_p$ and $\mathcal{S}_g$ denote predicted and ground-truth entity sets, and let $w(s)$ denote the geometric weight of an entity (triangle area for faces, edge length for edges). Weighted intersection and union are defined as $W_{\text{int}} = \sum_{s \in \mathcal{S}_p \cap \mathcal{S}_g} w(s), W_{\text{union}} = \sum_{s \in \mathcal{S}_p \cup \mathcal{S}_g} w(s)$.
Further we define metrics such as Intersection over Union (IoU), Precision and Recall on the mesh as follows: $\text{IoU} = \frac{W_{\text{int}}}{W_{\text{union}}},    \text{Precision} = \frac{W_{\text{int}}}{\sum_{s \in \mathcal{S}_p} w(s)},  \text{Recall} = \frac{W_{\text{int}}}{\sum_{s \in \mathcal{S}_g} w(s)}$.
This unified formulation applies to both faces and edges by simply modifying the geometric weight definition. The use of area and length weighting ensures that evaluation reflects true geometric coverage and structural correctness.

\begin{figure}[t]
    \centering
    \begin{subfigure}[b]{0.48\textwidth}
        \centering
        \includegraphics[width=\linewidth]{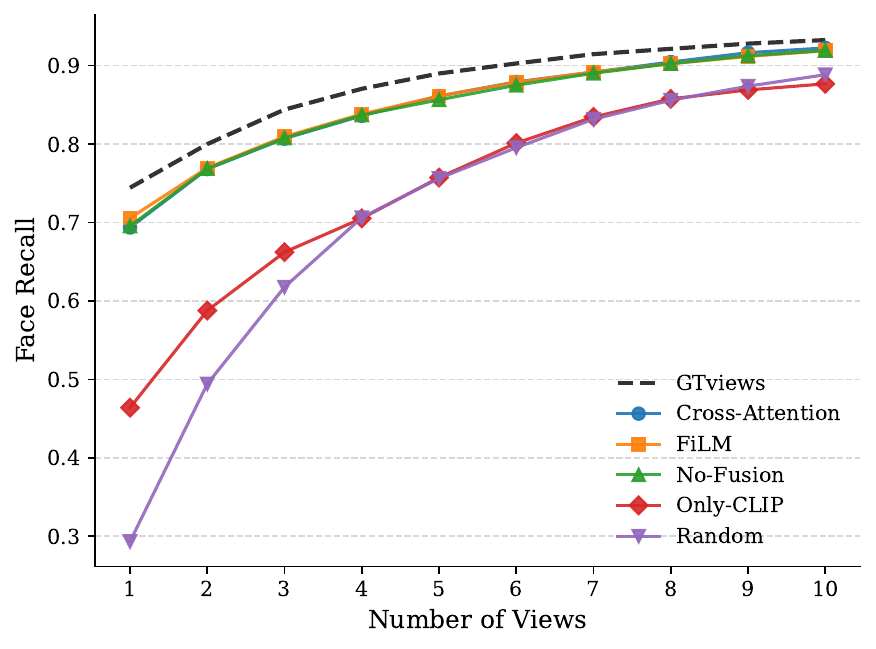}
        \caption{Face Recall}
        \label{fig:face_recall}
    \end{subfigure}
    \hfill
    \begin{subfigure}[b]{0.48\textwidth}
        \centering
        \includegraphics[width=\linewidth]{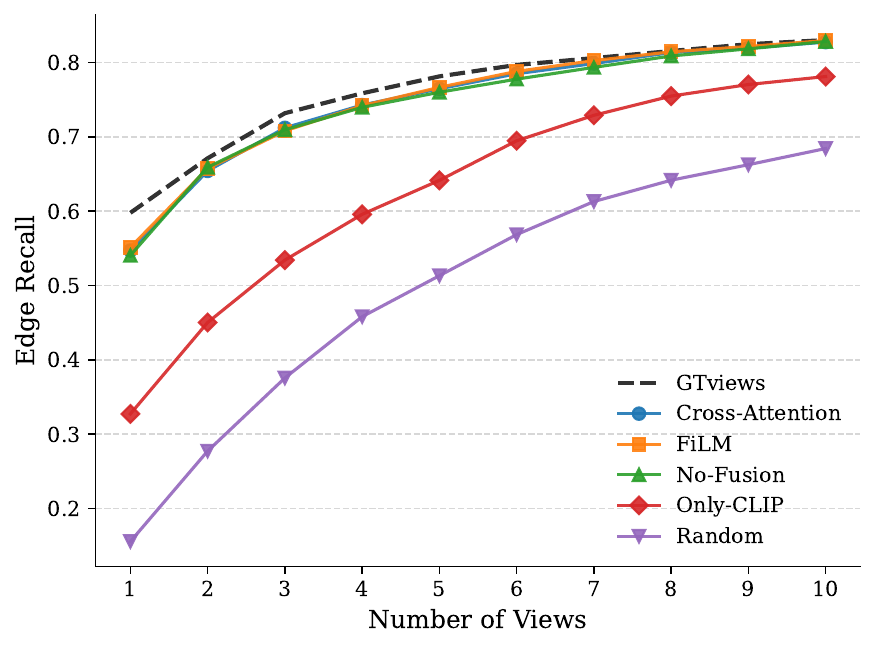}
        \caption{Edge Recall}
        \label{fig:edge_recall}
    \end{subfigure}
    \caption{Impact of view quantity on recall performance for Faces (a) and Edges (b) using CAD adapted LISA.}
    \label{fig:view_ablation}
\end{figure}

\begin{table}[h]
\centering
\caption{Performance evaluation of domain adapted and unadapted LISA variants and view selectors across Face@ top1 views and Edge@ top1 views metrics. \textit{GTviews}, short for the Ground Truth views, act as the oracle target views of CAD meshes on the test set: the single annotated view each entity was marked on and to which it's referring caption corresponds. Test set of 1535 queries (655 faces, 880 edges) from 218 meshes.}
\label{tab:lisa_view_selectors}
\resizebox{\linewidth}{!}{%
\setlength{\tabcolsep}{4pt}
\begin{tabular}{c >{\raggedright\arraybackslash}p{2.8cm} l cccc cccc}
\toprule
 & & & \multicolumn{4}{c}{\textbf{Face@ top1 views}} & \multicolumn{4}{c}{\textbf{Edge@ top1 views}} \\
\cmidrule(lr){4-7} \cmidrule(lr){8-11}
\textbf{S.no} & \textbf{Seg. VLM} & \textbf{View Selector} & \textbf{IoU} & \textbf{Prec.} & \textbf{Rec.} & \textbf{F1} & \textbf{IoU} & \textbf{Prec.} & \textbf{Rec.} & \textbf{F1} \\
\midrule

1 & LISA (Vanilla) & Cross-Attention & 0.292 & 0.337 & \textbf{0.652} & 0.390 & 0.043 & 0.048 & 0.494 & 0.074 \\
2 & LISA (Vanilla) & FiLM & \textbf{0.293} & \textbf{0.342} & 0.647 & \textbf{0.391} & \textbf{0.046} & \textbf{0.053} & \textbf{0.503} & \textbf{0.079} \\
3 & LISA (Vanilla) & No-Fusion & 0.293 & 0.338 & 0.647 & 0.389 & 0.044 & 0.049 & 0.502 & 0.077 \\
4 & LISA (Vanilla) & Only-CLIP & 0.201 & 0.239 & 0.460 & 0.272 & 0.035 & 0.039 & 0.388 & 0.061 \\
5 & LISA (Vanilla) & Random & 0.122 & 0.159 & 0.287 & 0.174 & 0.028 & 0.034 & 0.261 & 0.046 \\
6 & LISA (Vanilla) & \textit{GTviews} & \textit{0.293} & \textit{0.340} & \textit{0.645} & \textit{0.389} & \textit{0.046} & \textit{0.051} & \textit{0.510} & \textit{0.079} \\

\midrule

7 & LISA-CAD & Cross-Attention & 0.486 & 0.570 & 0.694 & 0.582 & \textbf{0.287} & \textbf{0.327} & 0.547 & \textbf{0.359} \\
8 & LISA-CAD & FiLM & \textbf{0.501} & \textbf{0.588} & \textbf{0.706} & \textbf{0.598} & 0.284 & 0.326 & \textbf{0.551} & 0.358 \\
9 & LISA-CAD & No-Fusion & 0.490 & 0.570 & 0.695 & 0.584 & 0.283 & 0.324 & 0.541 & 0.355 \\
10 & LISA-CAD & Only-CLIP & 0.305 & 0.374 & 0.464 & 0.377 & 0.160 & 0.193 & 0.327 & 0.210 \\
11 & LISA-CAD & Random & 0.196 & 0.262 & 0.293 & 0.250 & 0.071 & 0.097 & 0.156 & 0.101 \\
12 & LISA-CAD & \textit{GTviews} & \textit{0.529} & \textit{0.619} & \textit{0.744} & \textit{0.632} & \textit{0.311} & \textit{0.355} & \textit{0.598} & \textit{0.390} \\

\bottomrule
\end{tabular}%
}
\vspace{1mm}
\end{table}

\begin{table}[h]
\centering
\caption{\textbf{Comparison against zero-shot point cloud baselines} on the common 1535-query testset across 218 meshes. Existing point-cloud localization methods achieve high recall but low precision, resulting in poor F1 scores due to over-segmentation.}
\label{tab:baselines}
\resizebox{\linewidth}{!}{%
\setlength{\tabcolsep}{4pt}
\begin{tabular}{c l cccc cccc}
\toprule
& & \multicolumn{4}{c}{\textbf{Face Localization}} & \multicolumn{4}{c}{\textbf{Edge Localization}} \\
\cmidrule(lr){3-6}\cmidrule(lr){7-10}
\textbf{S.no} & \textbf{Model} & \textbf{IoU} & \textbf{Prec.} & \textbf{Rec.} & \textbf{F1} & \textbf{IoU} & \textbf{Prec.} & \textbf{Rec.} & \textbf{F1} \\
\midrule
1 & PartSLIP (Default)      & 0.073 & 0.083 & 0.368 & 0.114 & 0.012 & 0.013 & 0.422 & 0.022 \\
2 & PartSLIP (Top-PCD)      & 0.062 & 0.081 & 0.250 & 0.097 & 0.012 & 0.013 & 0.262 & 0.021 \\
3 & Find3D (Default)        & 0.171 & 0.172 & \textbf{0.955} & 0.260 & 0.017 & 0.017 & \textbf{0.937} & 0.032 \\
4 & Find3D (Top-PCD)        & 0.162 & 0.187 & 0.555 & 0.243 & 0.022 & 0.023 & 0.399 & 0.040 \\
5 & PatchAlign3D (Default)  & 0.159 & 0.161 & 0.893 & 0.242 & 0.017 & 0.017 & 0.895 & 0.031 \\
6 & PatchAlign3D (Top-PCD)  & 0.154 & 0.203 & 0.374 & 0.229 & 0.022 & 0.028 & 0.178 & 0.040 \\
\midrule
7 & MV-GEL (FiLM@top3, ViewNMS)        & 0.372 & 0.400 & 0.846 & 0.495 & 0.180 & 0.194 & 0.695 & 0.266 \\
8 & MV-GEL (FiLM@top3)                 & 0.419 & 0.469 & 0.810 & 0.540 & 0.228 & 0.248 & 0.708 & 0.317 \\
9 & \textbf{MV-GEL (FiLM@top1)} & \textbf{0.501} & \textbf{0.588} & 0.706 & \textbf{0.598} & \textbf{0.284} & \textbf{0.326} & 0.551 & \textbf{0.358} \\
\bottomrule
\end{tabular}%
}
\end{table}

\begin{figure}[t]
\centering

\begin{minipage}[t]{0.5\columnwidth}
\centering
\vspace{0pt}

\includegraphics[
    width=\linewidth,
]{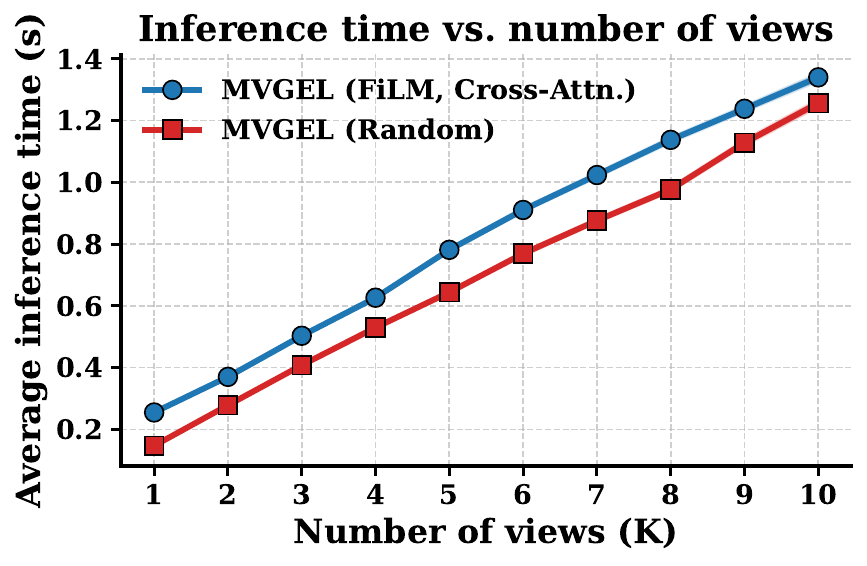}

\end{minipage}
\hfill
\begin{minipage}[t]{0.44\columnwidth}
\centering
\vspace{0pt}

\begin{minipage}{0.48\linewidth}
    \includegraphics[width=0.8\linewidth]
    {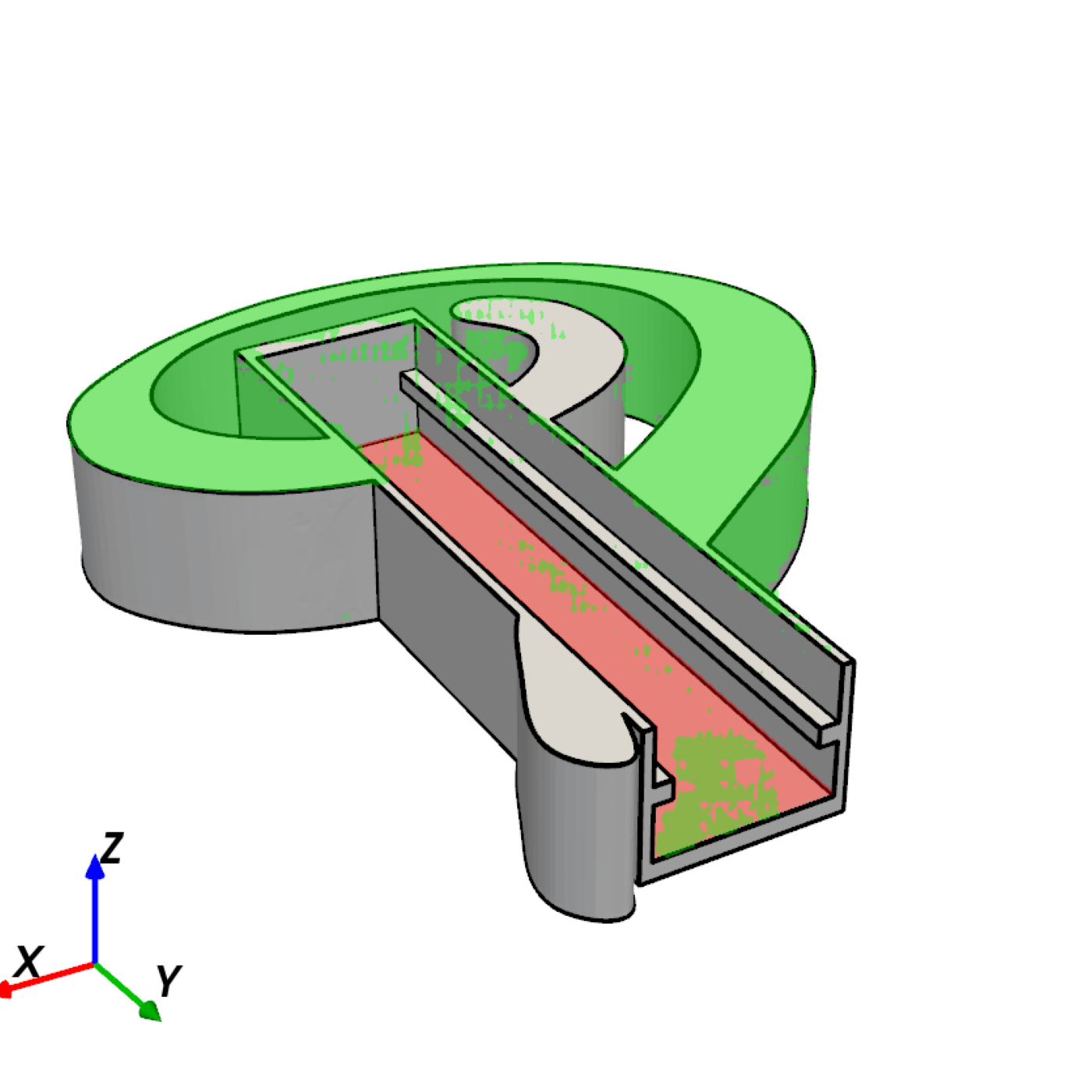}
\end{minipage}
\hfill
\begin{minipage}{0.48\linewidth}
    \includegraphics[width=0.8\linewidth]
    {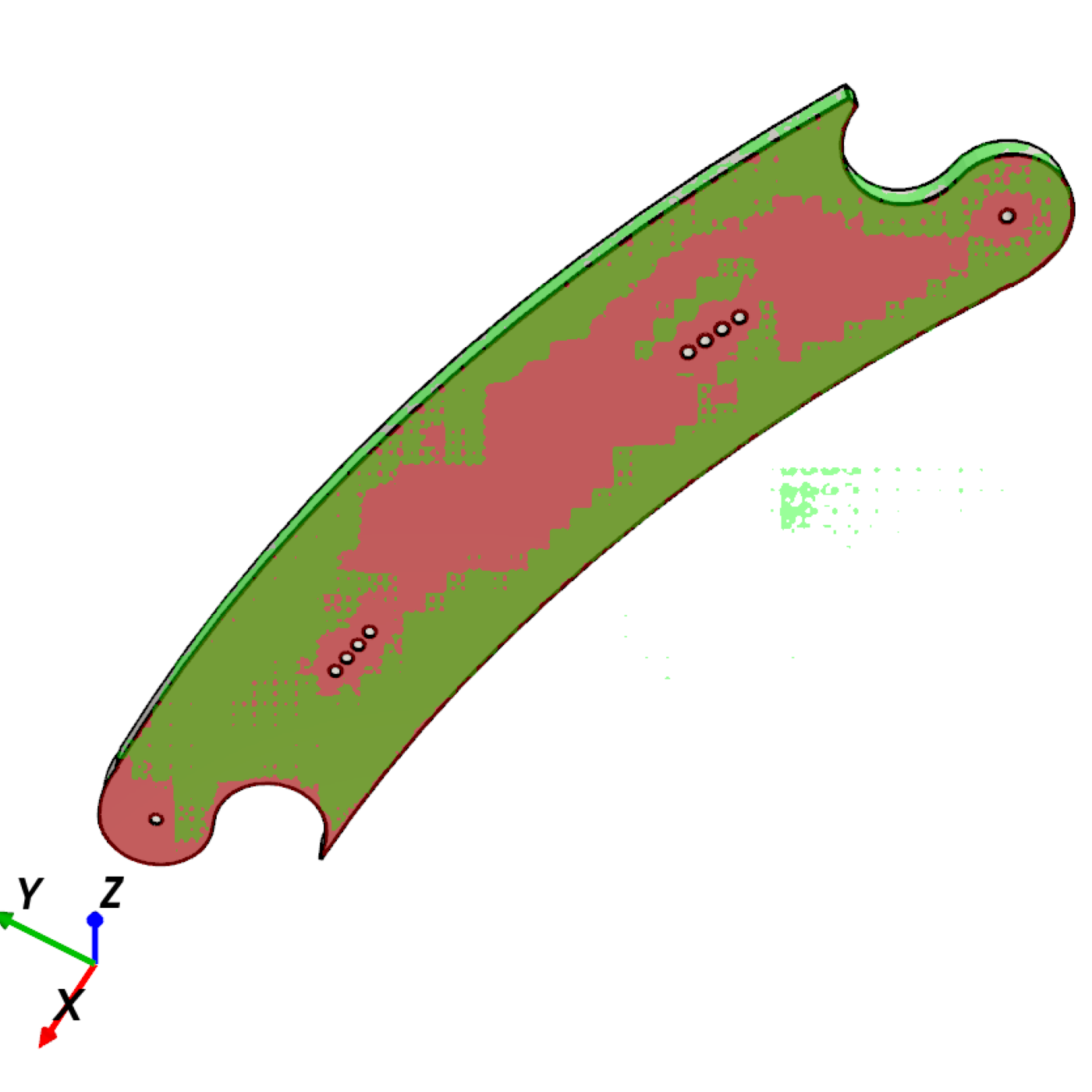}
\end{minipage}

\vspace{1mm}

\begin{minipage}{0.48\linewidth}
    \includegraphics[width=0.8\linewidth]
    {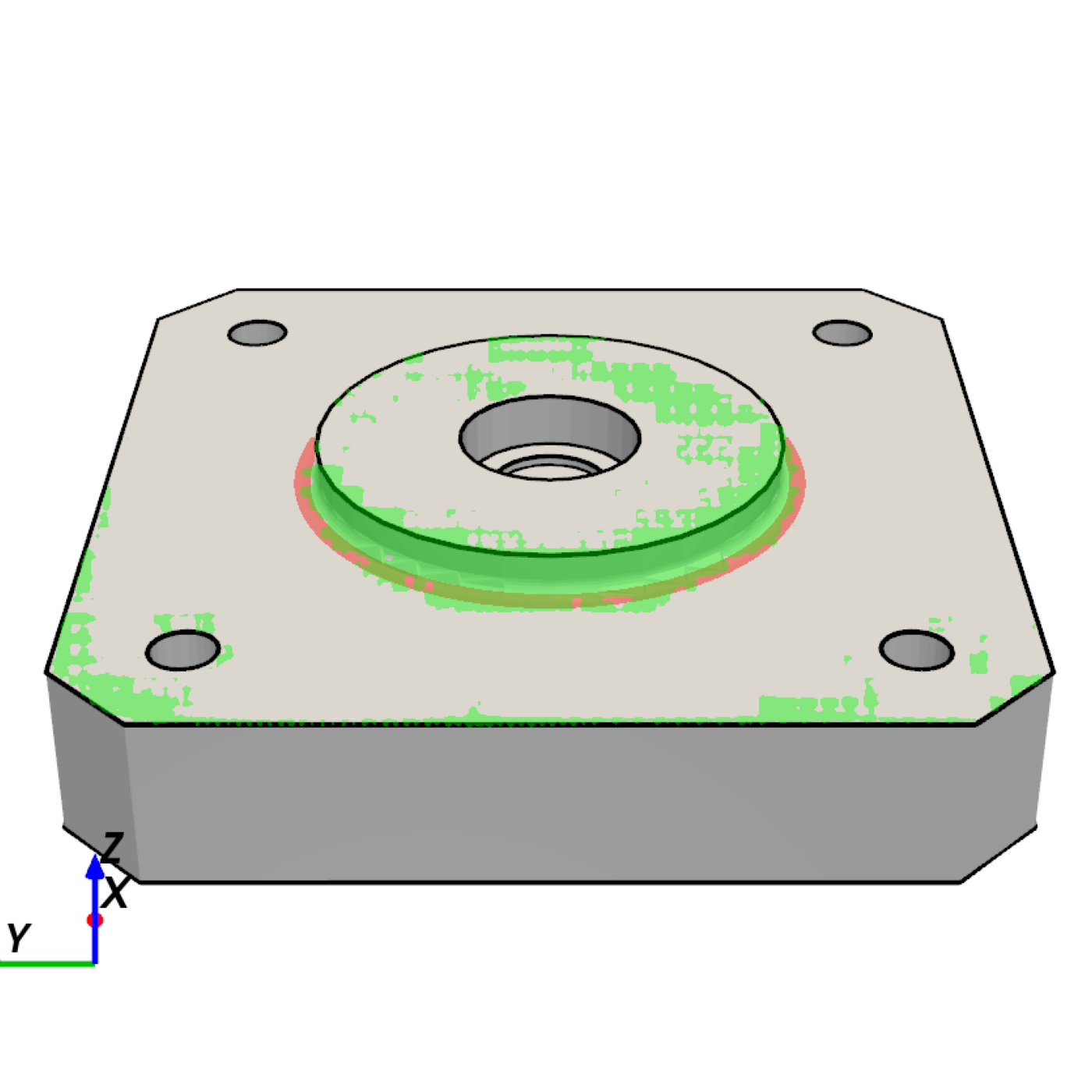}
\end{minipage}
\hfill
\begin{minipage}{0.48\linewidth}
    \includegraphics[width=0.8\linewidth]
    {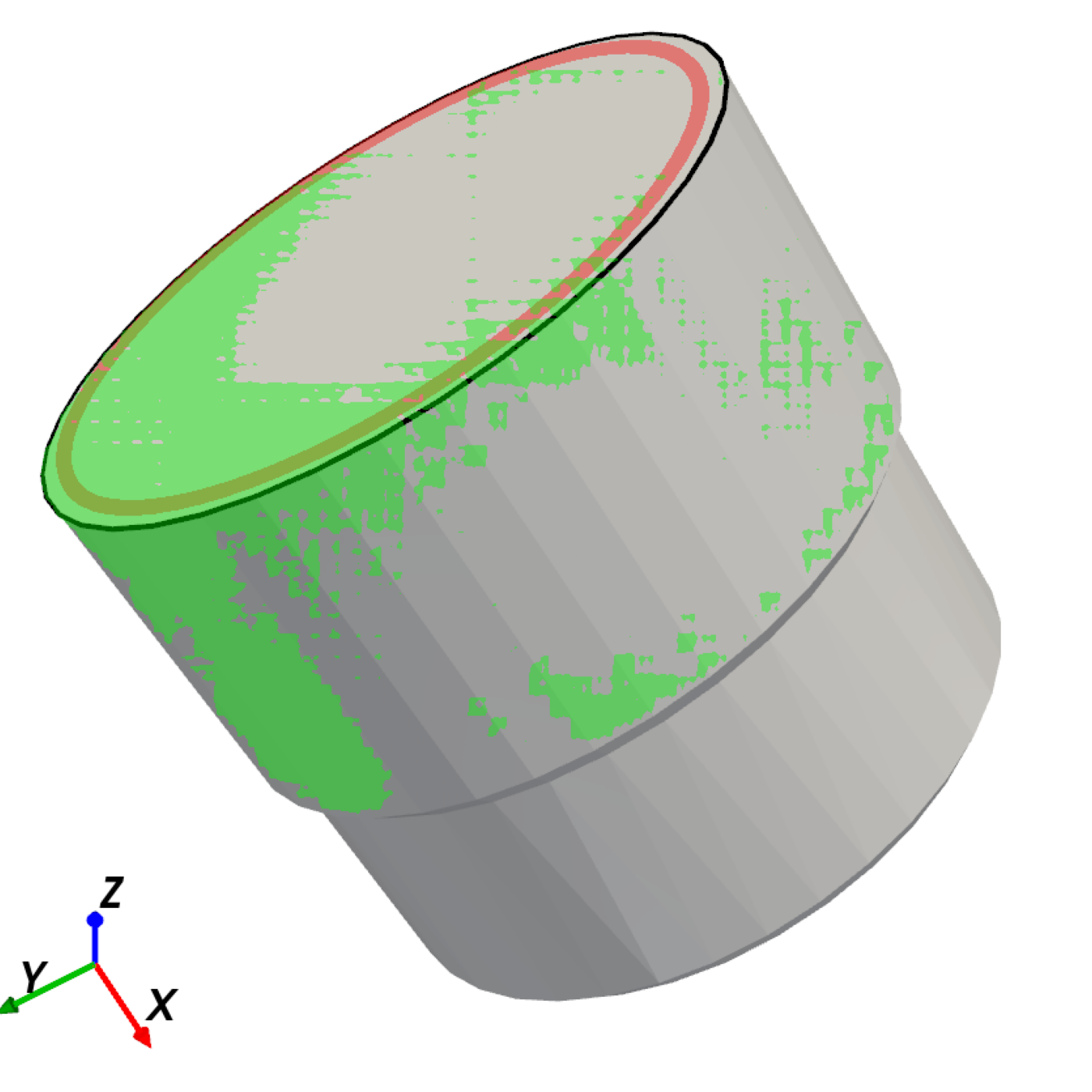}
\end{minipage}

\end{minipage}

\caption{
\textbf{Inference efficiency and qualitative localization results.}
\textbf{Left:} Inference latency as a function of the number of selected views ($K$). The proposed ranking module adds only $\sim$0.1\,s ranking module overhead while achieving comparable face recall using a single selected view, whereas random selection requires at least five views.
\textbf{Right:} Representative LISA-Vanilla localizations. The top row shows face localization examples and the bottom row shows edge localization examples. Red regions denote ground-truth target segments and green regions denote predicted segments.
}
\label{fig:latency_and_qualitative}
\end{figure}

\subsubsection{Ablation Studies}

We analyze the impact of view ranking formulation and supervision granularity to understand the role of geometric prioritization in language-driven CAD localization. Using the geometric metrics defined earlier, we evaluate segmentation adaptation and view selection design. The metrics reported were averaged for all samples in the test set. Results for Face@top1 view and Edge@top1 view are shown in Table~\ref{tab:lisa_view_selectors}.

We ablate components of \textit{GELviews}, comparing FiLM and Cross-Attention fusion variants applied in the modality fusion block (Fig.\ref{fig:GELviews}). The No-Fusion variant bypasses this block by directly using position-aware features for self attention in the view-interaction head, while Only-CLIP removes the view-interaction head and relies solely on the CLIP head. Random view selection is included to assess the net contribution of learned ranking. All view selection strategies are evaluated with both fine-tuned (LISA-CAD) and Vanilla LISA backbones. For a better understanding of the ablated networks, readers are advised to refer to the supplementary.

\paragraph{Effect of Domain Adaptation}

As seen in Table \ref{tab:lisa_view_selectors},  replacing Vanilla LISA with LISA-CAD consistently improves performance across all view selectors for both faces and edges. While Vanilla LISA attains moderate recall, it exhibit's low precision on edges and faces, leading to weak IoU and F1, as shown in Fig. \ref{fig:latency_and_qualitative}(right) . This indicates over-segmentation and limited geometric specificity. CAD fine-tuning substantially improves precision while preserving recall, resulting in markedly stronger F1 and IoU. These results confirm that domain adaptation is critical for geometry-aware segmentation.

\paragraph{Effect of View Selector Design}

Within CAD-LISA, fusion-based view selectors outperform Only-CLIP and Random view selection strategies and perform on-par with No-Fusion. FiLM achieves consistently strong overall performance across top views (for top@3 and top@5 views please refer to the supplementary), closely followed by Cross-Attention and No-Fusion. Cross-Attention yields better metrics for edge localization. Cross-Attention, despite being a powerful multimodal fusion architecture, doesn't outperform FiLM by a great margin. We provide a reason for this performance order in the supplementary. The consistently minor win of fusion-based models over No-Fusion variant indicates that interaction between semantic and view-aware features improves ranking quality. The Only-CLIP selector performs significantly worse, particularly for edges, demonstrating that global image-text similarity alone is insufficient for identifying geometrically informative viewpoints. Random view selection yields the lowest performance across both backbones, confirming that gains arise from learned view prioritization. Edge localization is consistently more challenging than face localization due to thin projections, discretization sensitivity, and view-dependent occlusions. Incorrect views can prevent the segmentation backbone from observing the relevant entities described in $q$, leading to missed detections. Nevertheless, relative improvements across selectors are larger for edges, highlighting the importance of view quality and domain-specific segmentation for structurally sensitive features. The combination of CAD-LISA and fusion-based ranking provides the most balanced performance across both entity types.



\paragraph{Visibility Oracle and View-Selection Headroom}

To quantify the absolute effectiveness of our view selection module, we introduce a \textit{GTviews} Ground truth oracle baseline in Table \ref{tab:lisa_view_selectors}. This acts as the theoretical performance ceiling, representing the maximum achievable scores when the target geometric view aligned most with the text prompt is selected. Remarkably, our best-performing FiLM selector achieves a Face F1 score of 0.6, capturing ~95\% of the oracle’s performance limit (0.63 F1). Similarly, for the highly challenging task of edge localization, FiLM reaches an F1 of 0.36, realizing more than 90\% of the 0.39 F1 ceiling. This minimal performance gap proves that our prompt-aware view-ranking network operates near-optimally in identifying the best possible viewpoint, bottlenecked only by the inherent segmentation limit's of the reasoning backbone itself. Readers are advised to refer to the supplementary for additional tables with metric analysis for higher top-k values.

\paragraph{Multi-View Aggregation}

Top-K recall curves, as shown in Fig.~\ref{fig:view_ablation}, further validate the ranking quality and view diversity of our proposed modules. The \textit{(GTviews)} trajectory serves as the upper bound for multi-view aggregation. Notably FiLM, Cross-Attention and No-Fusion rapidly close the gap to this oracle within the first 7 to 8 views. The high initial recall demonstrates that the network successfully prioritizes the most geometrically informative views first. 

In contrast, the Only-CLIP and Random LISA-CAD baselines start at significantly lower recall and require a large number of views to gradually accumulate surface coverage, proving that purely semantic or arbitrary sampling fails to capture view-dependent geometry efficiently. This advantage is especially pronounced for edge localization (Figure~\ref{fig:view_ablation}b), where geometry-aware fusion dramatically accelerates the coverage of thin, easily occluded entities. 

\paragraph{Baseline Comparisons}
In Table \ref{tab:baselines}, we evaluated similar open vocabulary point cloud based segmentation baselines like PartSLIP \cite{liu2023partslip}, Find3D \cite{ma2025find}, PatchAlign3D \cite{hadgi2026patchalign3d} on our test dataset, by projecting point predictions back to the mesh. We introduced default as well as adapted \textit{Top-PCD} (restricting to top point cloud density predictions matching the caption: 10\% for faces, 2\% for edges) baselines. Because these models work with point clouds that lack explicit boundary topology and lack deep language reasoning, they severely over-segment (``bleed'') across faces (\eg, Find3D achieves 0.95 Recall but only 0.17 Precision) and 1D edges (IoU $\le$ 0.022).

\paragraph{Inference Cost vs. Views}
Fig. \ref{fig:latency_and_qualitative}(Left)  explicitly profiles the avg. time/query vs. views. Reaching ~0.7 recall on faces (on average on the test set) with 5 random views incurs $>2\times$ the latency of 1 \textit{GELview} (0.6s for the random views variant vs 0.25s for the FiLM variant). Using \textbf{MV-GEL} random takes ~16 GB VRAM while MV-GEL (FiLM) takes ~17GB for one query on average. Thus \textit{GELviews} yields significantly better metrics at a fraction of the computational cost. The inference was profiled on a single H100 NVIDIA GPU.



\begin{figure}[t]
    \centering
    \setlength{\abovecaptionskip}{4pt}
    \setlength{\belowcaptionskip}{0pt}

    \begin{subfigure}{0.31\linewidth}
        \centering
        \includegraphics[height=3cm]{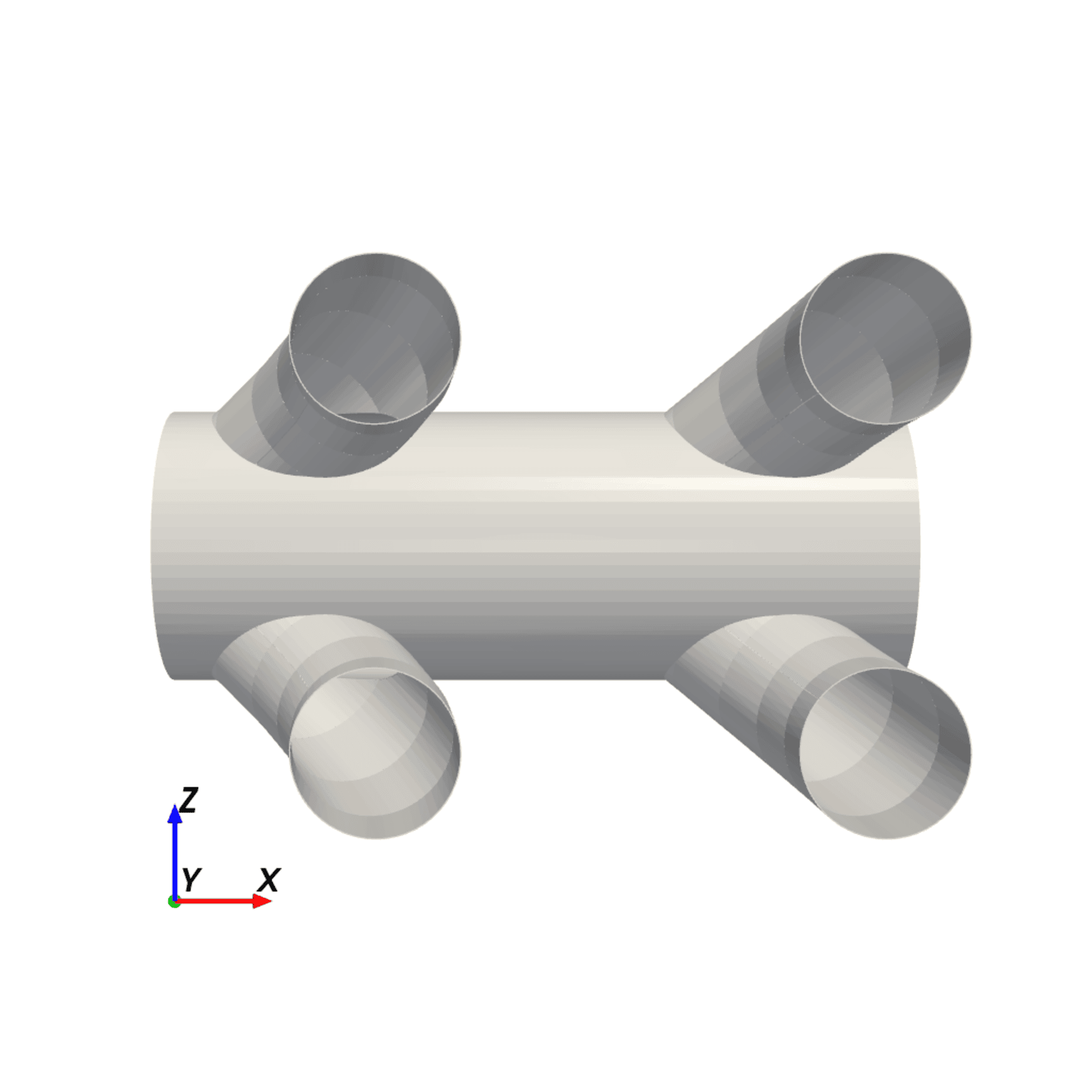}
    \end{subfigure}
    \hfill
    \begin{subfigure}{0.31\linewidth}
        \centering
        \includegraphics[height=3cm]{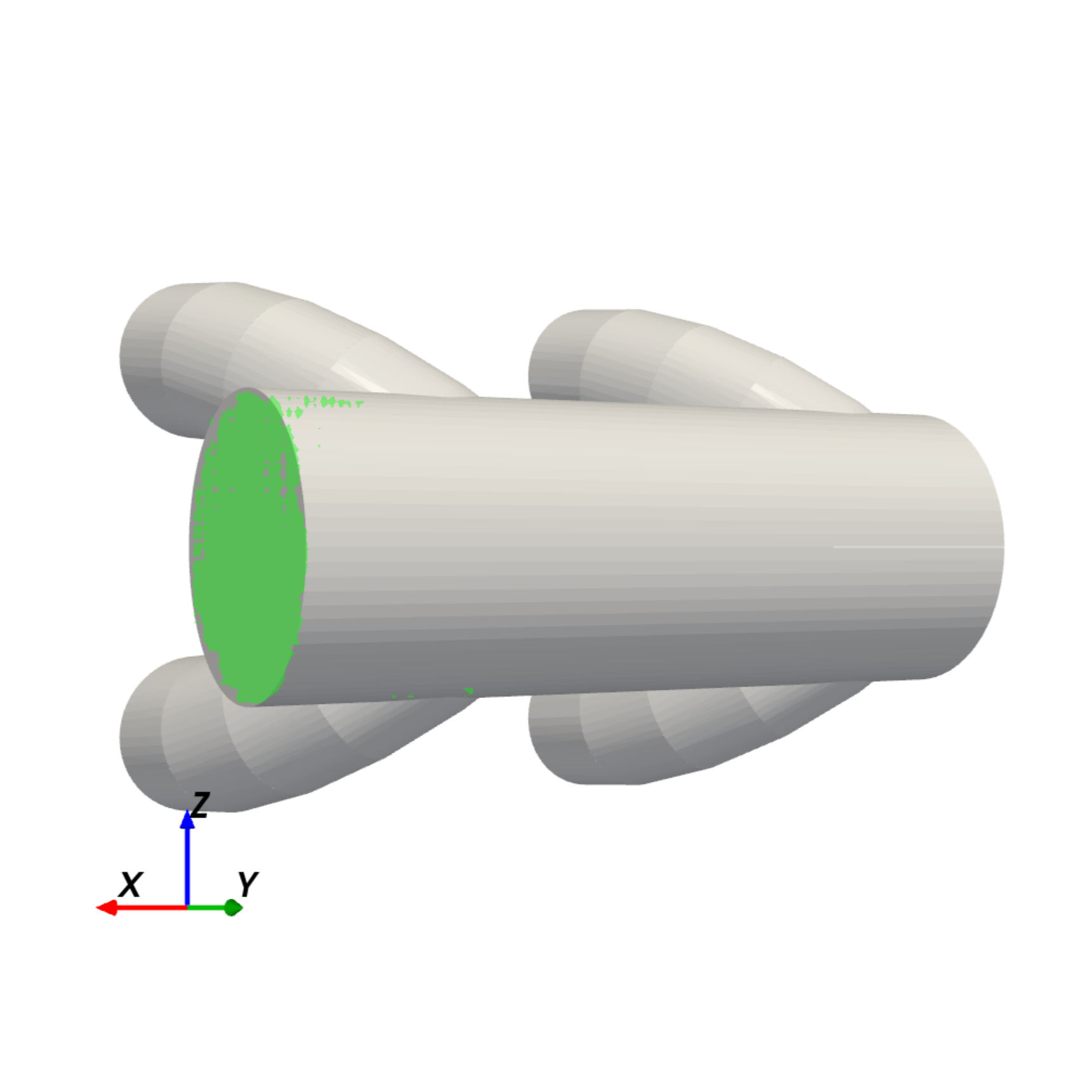}
    \end{subfigure}
    \hfill
    \begin{subfigure}{0.31\linewidth}
        \centering
        \includegraphics[height=3cm]{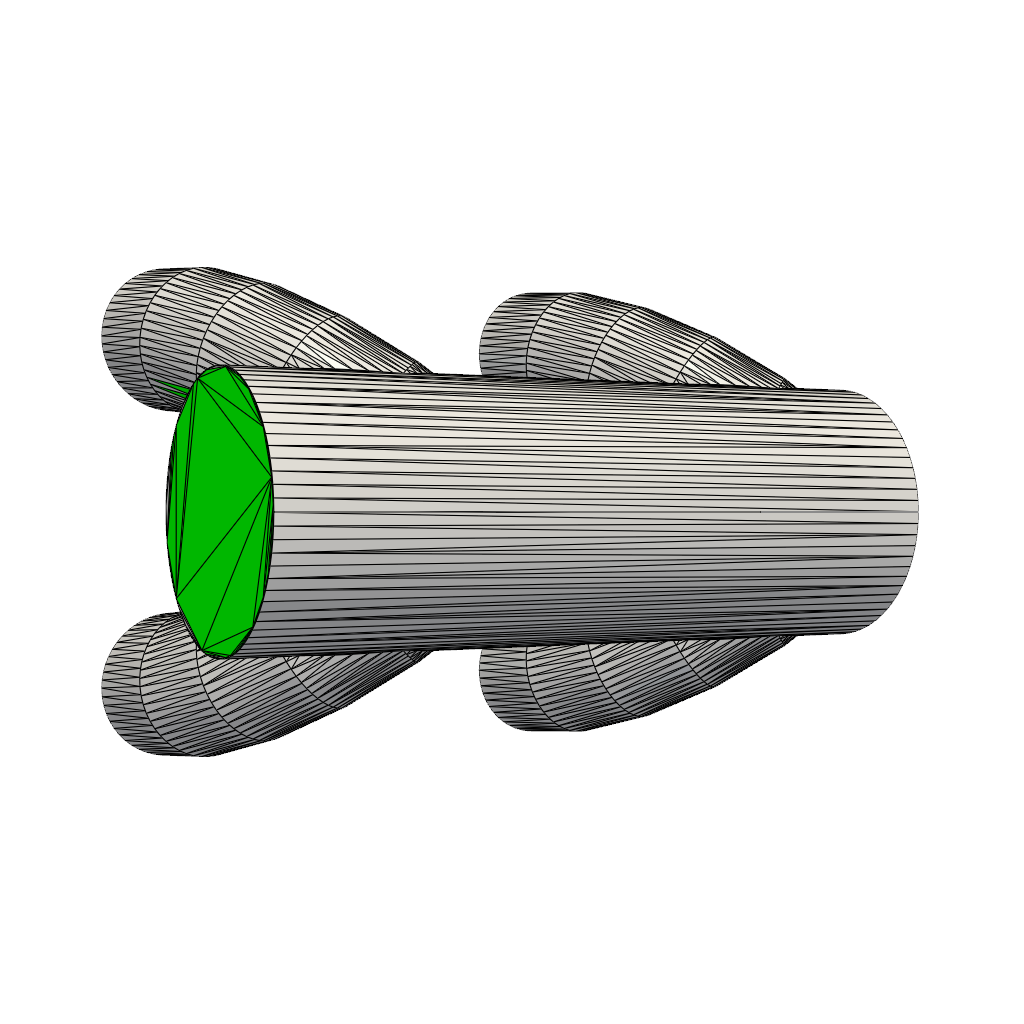} 
    \end{subfigure}

    \vspace{-1mm}
    \caption*{\footnotesize \textbf{Caption}: It's the large flat circular face at one end of the pipe assembly. It's best seen from the right end of the pipe at a level angle.}

    \begin{subfigure}{0.31\linewidth}
        \centering
        \includegraphics[height=3cm]{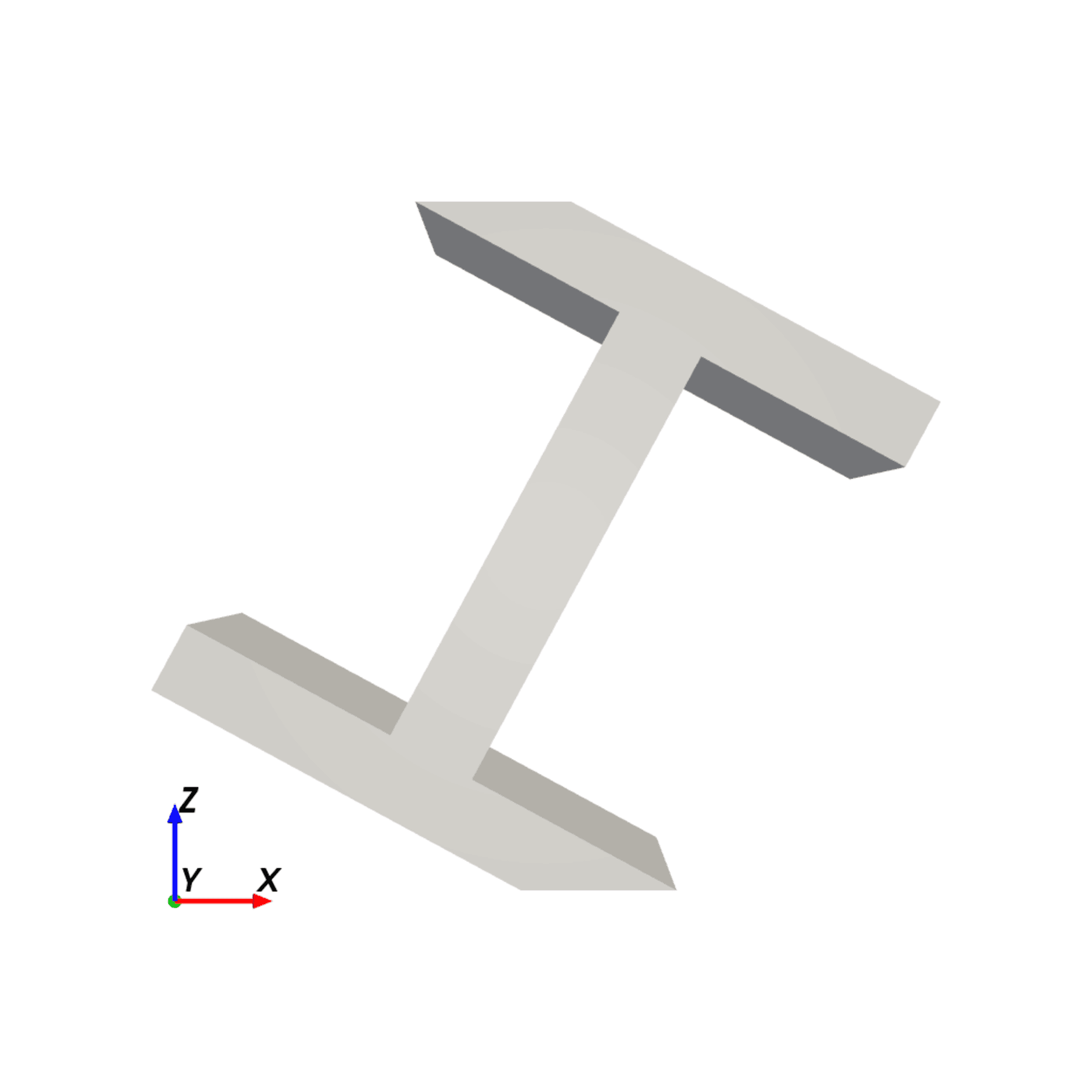}
    \end{subfigure}
    \hfill
    \begin{subfigure}{0.31\linewidth}
        \centering
        \includegraphics[height=3cm]{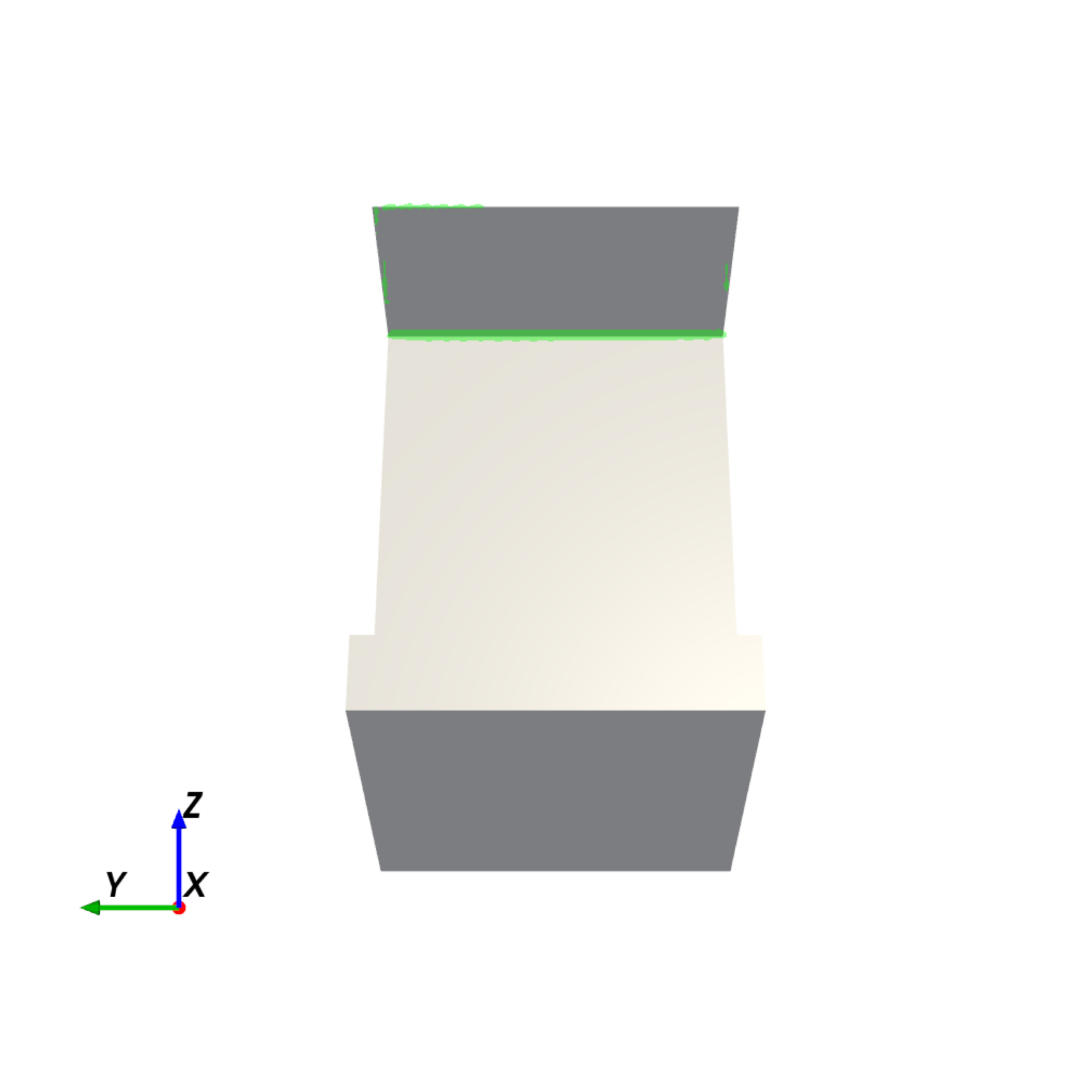}
    \end{subfigure}
    \hfill
    \begin{subfigure}{0.31\linewidth}
        \centering
        \includegraphics[height=3cm]{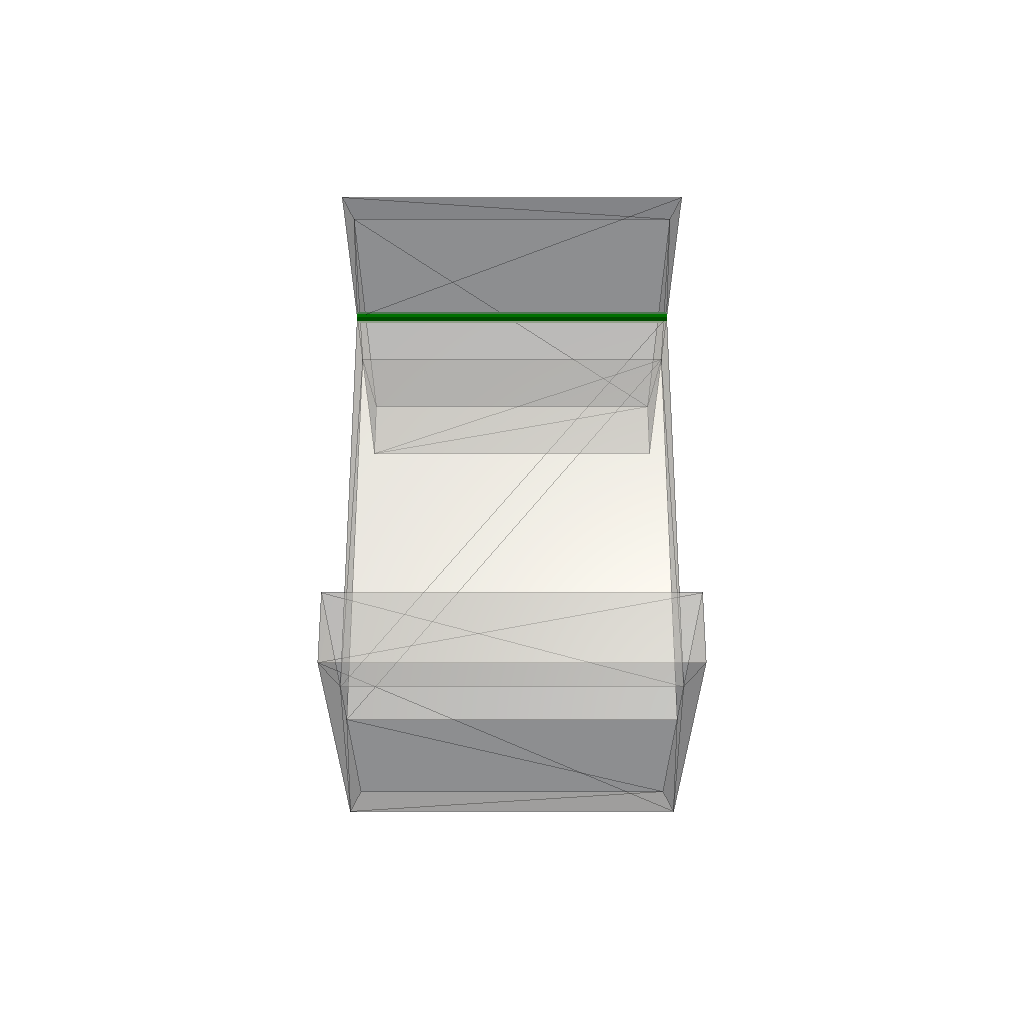} 
    \end{subfigure}

    \vspace{-1mm}
    \caption*{\footnotesize \textbf{Caption}: It's the internal corner edge around the top of the I- structure when seen from the left, looking from underneath.}

    \vspace{2mm}
    \caption{\footnotesize \textbf{MV-GEL} localization pipeline. (Left) Default mesh view, (Center) \textit{GELviews} selected view (top-1) segmented by LISA-CAD, (Right) MV-GEL localized entity predictions on the mesh.}
    \label{fig:my_3x2_figure}
\end{figure}

\section{Conclusions}

We introduced \textbf{MV-GEL}, a framework for language-driven geometric entity localization that unifies reasoning-based segmentation with prompt-aware view selection. By adapting LISA to the CAD domain and proposing \textit{GELviews}—a geometry-conditioned view ranking module, our method accurately localizes specific target faces and edges without requiring parametric B-Rep data.

Our extensive ablations reveal two critical insights. First, domain-specific fine-tuning is essential; it resolves the severe over-segmentation of thin structural edges seen in open-domain models. Second, our fusion-based view selection dramatically outperforms semantic-only (CLIP) and random sampling. By actively prioritizing geometrically informative viewpoints, our method captures over 90\% of the theoretical oracle performance for top-1 face and edge localization, and rapidly accelerates surface coverage during multi-view aggregation.

Ultimately, our results demonstrate that language-driven 3D localization is fundamentally a geometry-aware view prioritization problem, not merely an image-text alignment task. By bridging multimodal reasoning with object understanding from the views, \textbf{MV-GEL} establishes an effective solution for geometric entity localization and understanding.


\section*{Acknowledgements}
The authors gratefully acknowledge Helmholtz-Zentrum Hereon for providing the
GPU computing resources on which all experiments in this work were carried out, and thank Prof.\ Christian Cyron, Head of the Institute of Material Systems Modeling (MSM), Helmholtz-Zentrum Hereon, for providing the platform and support that made this work possible. The authors also thank their colleagues at MSM for helpful discussions. The authors have no competing interests to declare that are relevant to the content of this article.

\clearpage
%
%
\bibliographystyle{splncs04}
\bibliography{main}

\clearpage



\end{document}